\documentclass[sigconf]{aamas}

\usepackage{balance}
\usepackage{algorithm}
\usepackage{algpseudocode}
\usepackage{subfigure}
\usepackage{changepage}
\usepackage{pdfpages}

\doi{DCBH3506}

\setcopyright{ifaamas}
\makeatletter
\@printcopyrighttrue
\makeatother
\acmConference[AAMAS '26]{Proc.\@ of the 25th International Conference
on Autonomous Agents and Multiagent Systems (AAMAS 2026)}{May 25 -- 29, 2026}
{Paphos, Cyprus}{C.~Amato, L.~Dennis, V.~Mascardi, J.~Thangarajah (eds.)}
\copyrightyear{2026}
\acmYear{2026}
\doi{DCBH3506}
\acmPrice{}
\acmISBN{}

\acmSubmissionID{<<144>>}

\title[AAMAS-2026 Formatting Instructions]{Repeated Deceptive Path Planning against Learnable Observer }

\author{Shiyue Cao$^{1,2}$, Pei Xu$^{2}$, Likun Yang$^{1,2}$, Lei Cui$^{1,2}$, Shizhao Yu$^{1,2}$, Shiyu Zhang$^{2}$, Yongjian Ren$^{1,2}$, Xiaotang Chen$^{2}$, Kaiqi Huang$^{1,2}$}

\affiliation{
  \institution{$^1$School of Artificial Intelligence, University of Chinese Academy of Sciences}
  \city{Beijing} \country{China}
}
\affiliation{
  \institution{$^2$National Key Laboratory of Cognition and Decision Intelligence for Complex Systems, Institution of Automation, Chinese Academy of Sciences}
  \city{Beijing} \country{China}
}

\email{{caoshiyue2021, pei.xu, yanglikun2021, cuilei2024, yushizhao2022, shiyu.zhang, renyongjian2022}@ia.ac.cn}
\email{{xtchen, kaiqi.huang}@nlpr.ia.ac.cn}

\begin{abstract}
We study the problem of deceptive path planning (DPP), where an agent aims to conceal its true destination from external observers. While existing work assumes static, non-learning observers, real-world adversaries—such as in critical goods transportation or military operations—can adapt by learning from historical trajectories.
To address this gap, we introduce Repeated Deceptive Path Planning (RDPP), a new formulation that explicitly models learnable observers. We show that existing DPP methods fail under this setting, as they cannot adapt to evolving adversarial predictions.
While incorporating observer previous predictions into updates enables some adaptation, such incremental updates cause accumulative lag that degrades deception.
To this end, we propose \textbf{De}ceptive \textbf{M}eta \textbf{P}lanning (DeMP), a two-level optimization framework that combines episode-level adaptation,  which enables short-term policy adjustment to counter updated observer, and meta-level updates, which leverage cross-episode feedback to capture how observers update their models and accelerate adaptation in future episodes.
In this way, DeMP mitigates the accumulation of adaptation lag, enabling sustained deception against a learning observer.
Experiments across environments demonstrate that DeMP significantly outperforms existing approaches in RDPP while maintaining competitive path cost.  Our results highlight the importance of modeling repeated interactions with learnable adversaries, providing new insights into deception and privacy in multi-agent systems.

\end{abstract}


\newcommand{\BibTeX}{\rm B\kern-.05em{\sc i\kern-.025em b}\kern-.08em\TeX}

\makeatletter
\gdef\@mkbibcitation{
  \par\medskip\small\noindent{\bfseries ACM Reference Format:}\par\nobreak
  \noindent\bgroup
    \def\\{\unskip{}, \ignorespaces}
    Shiyue Cao, Pei Xu, Likun Yang, Lei Cui, Shizhao Yu, Shiyu Zhang, Yongjian
    Ren, Xiaotang Chen, and Kaiqi Huang. 2026.
    Repeated Deceptive Path Planning against Learnable Observer : Extended Abstract.
    In Proc. of the 25th International Conference on Autonomous Agents and Multiagent Systems
    (AAMAS 2026), Paphos, Cyprus, May 25 -- 29, 2026, IFAAMAS, 3 pages.
    \url{https://doi.org/10.65109/DCBH3506}
  \par\egroup
}
\gdef\@copyrightpermission{
  \begin{minipage}{0.2\columnwidth}
   \href{https://creativecommons.org/licenses/by/4.0/}{\includegraphics[width=0.90\textwidth]{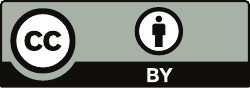}}
  \end{minipage}\hfill
  \begin{minipage}{0.8\columnwidth}
   \href{https://creativecommons.org/licenses/by/4.0/}{This work is licensed under a Creative Commons Attribution International 4.0 License.}
  \end{minipage}
  \par
  This is the full version of an accepted extended abstract. The official conference bibliographic
  information appears below.
  \vspace{5pt}
}
\makeatother

\begin{document}

\pagestyle{fancy}
\fancyhead{}

\maketitle

\section{Introduction}

\begin{figure}[h]
		\centering
 \includegraphics[width=0.9\linewidth]{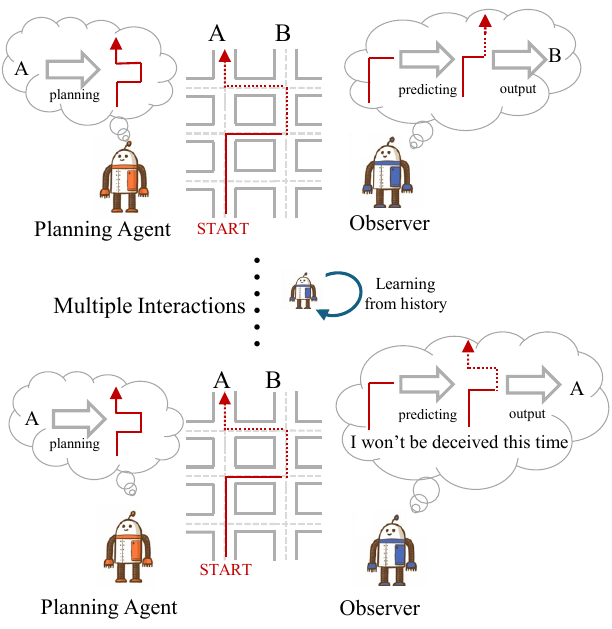}

		\caption{Illustration of Repeated Deceptive Path Planning (RDPP). Unlike single-shot DPP, RDPP introduces a learnable observer. After each episode, the observer receives the full trajectory of agent to update its recognition model, while the agent accesses the  predicted goal of observer.
        This highlights the core challenge of RDPP: an agent must adapt its policy to achieve sustained deception over multiple interactions.} \label{fig1}

\end{figure}

Privacy protection in planning is a critical concern in adversarial scenarios, where agents must achieve their goals while preventing external observers from inferring their true intentions. This challenge arises in many real-world applications~\cite{Bell2003TowardAT,xu2022path,luo2019opponent}. For example, when cash trucks transport money in urban environments, their routes may be observed by potential threats, making it necessary to design paths that conceal the actual destination.

\begin{figure*}[t]
		\centering
        \includegraphics[width=0.90\linewidth]{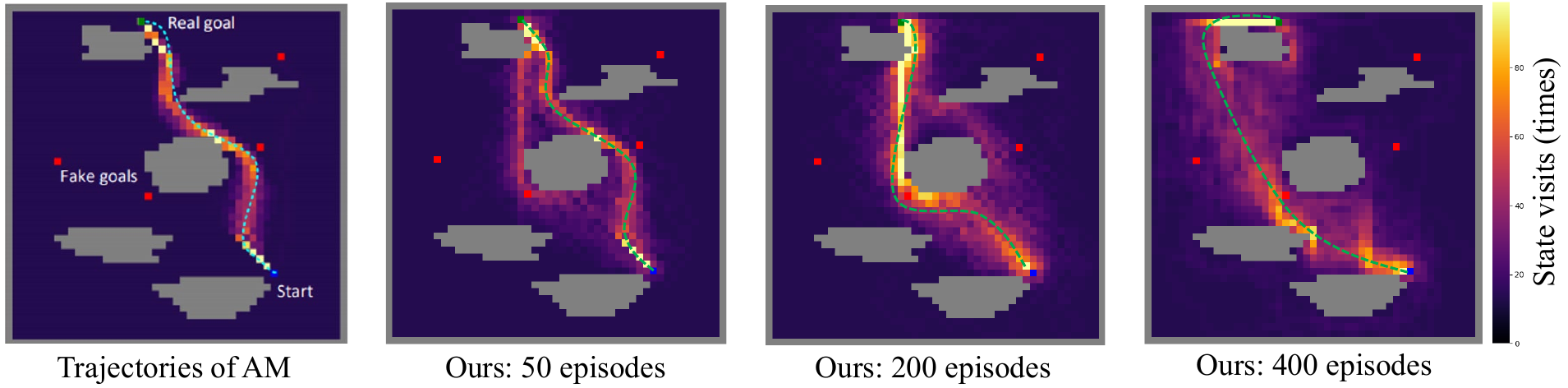}

		\caption{Trajectory Evolution of DeMP in RDPP. The left upper panel shows the static trajectory of the baseline method AM. The remaining panels are State Visits Heatmaps illustrating DeMP's path distribution over repeated interactions, where brighter colors indicate more frequent visits and the dashed line is the final path. Unlike AM, DeMP demonstrates continuous path evolution and diverse path adaption, successfully preventing the learnable observer from inferring the true goal based on a single  movement pattern.}
\label{heatfig}

\end{figure*}
Deceptive Path Planning (DPP)~\cite{masters_deceptive_2017} addresses this problem by generating trajectories that obscure the agent's true goal from an adversarial observer. Existing approaches span path planning~\cite{masters_deceptive_2017,savas_deceptive_2022}, control-based methods~\cite{wagner2011acting,dragan2014analysis,ornik2018deception}, objective optimization~\cite{gutierrez2025agent}, and reinforcement learning~\cite{liu_deceptive_2021,lewis_deceptive_2023,fatemi_deceptive_2024}. However, most of these methods focus on one-shot interactions and assume static, non-learnable observers. In practice, adversaries often monitor agents across repeated interactions—for example, in military operations, where opponents can analyze troop movements over extended periods—and continuously refine their inference strategies by exploiting historical trajectories. Under such repeated settings, the effectiveness of existing DPP methods deteriorates rapidly~\cite{ornik2018deception}, highlighting a fundamental limitation of one-shot deceptive planning.

Building on DPP, we introduce Repeated Deceptive Path Planning (RDPP), which extends deceptive planning to repeated interactions with a learnable observer. As illustrated in Figure~\ref{fig1}, in each episode the agent executes a trajectory toward its true goal while attempting to mislead the observer, who predicts the goal based on a partial trajectory prefix. After the episode, the observer updates its recognition model using the full trajectory and true goal, while the agent gains access to the observer's prediction as feedback for future interactions. This formulation captures a fundamental challenge absent in single-shot DPP: deception must be sustained against an observer whose recognition capabilities improves over time.

A natural approach for RDPP is to incorporate the observer's predictions into the reward and update the agent's policy after each episode. However, this reactive strategy suffers from a structural limitation. Because the agent adapts only after observing the observer's improvement, its policy updates consistently trail the observer's learning progress. Over repeated interactions, this mismatch leads to an accumulation of adaptation lag, causing previously effective deceptive behaviors to become predictable and eventually ineffective.

To address this issue, we propose \textbf{Deceptive Meta Planning (DeMP)}, a two-level optimization framework that combines episode-level adaptation with meta-level updates across episodes.
At the episode-level, the agent adapts its strategy after each episode of interaction to counter the observer's updated recognition model.
Crucially, at the meta-level, DeMP integrates long-term feedback using higher-order gradient information.
Rather than merely reacting to past updates, this mechanism mathematically anticipates the observer's learning dynamics. This enables proactive adaptation, allowing the agent to identify policy initializations that are robust to the observer's future decision boundary shifts.
By doing so, DeMP theoretically mitigates the accumulation of adaptation lag inherent in sequential adversarial interactions, enabling sustained deception.

We evaluate RDPP and DeMP in a set of grid-world environments with adaptive observers. Experimental results demonstrate that existing DPP methods suffer significant degradation under repeated interactions, whereas DeMP sustains high deception performance while maintaining competitive path costs. These findings highlight the effectiveness of our approach and emphasize the importance of considering learnable adversaries in repeated deceptive path planning scenarios.

Overall, we make three key contributions to deceptive planning:
\begin{itemize}
    \item We formalize Repeated Deceptive Path Planning (RDPP), extending standard DPP to repeated interactions with learnable observers.
    \item We propose Deceptive Meta Planning (DeMP), a two-level framework supported by theoretical analysis that justifies its surrogate objective alignment and validates its proactive adaptation against evolving observers.
    \item We empirically demonstrate that DeMP sustains high deception performance while maintaining competitive path efficiency in repeated interactions.
\end{itemize}

\begin{figure*}[h]
		\centering
        \includegraphics[width=0.9\linewidth]{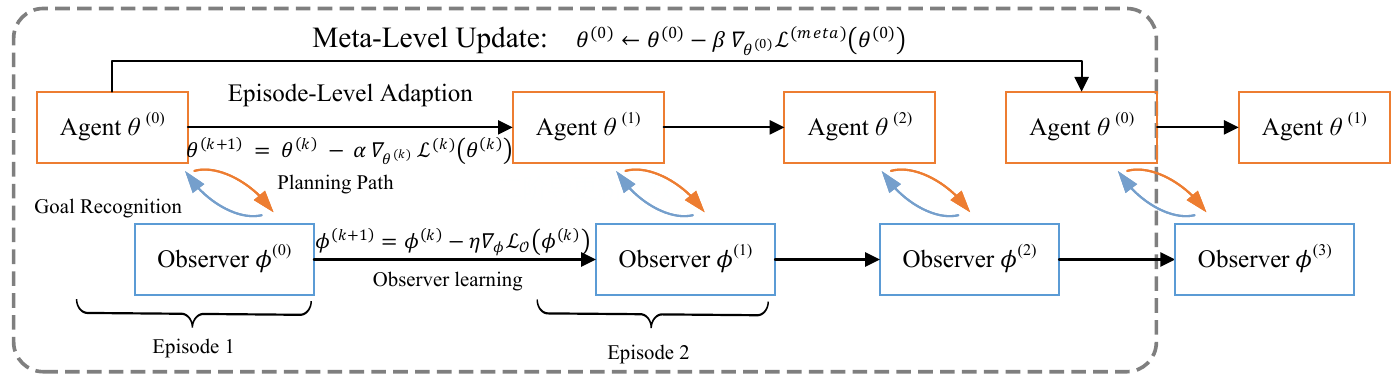}
		\caption{ The two-level optimization framework of DeMP. The process is structured into two levels: (1) The Episode-Level Adaptation involves $M$ episodes where the Agent adapts its policy parameter $\theta^{(k)}$ to counter the Observer's updated predictions. The Observer updates its recognition model after each episode. (2) The Meta-Level Update (dashed box) occurs after $M$ episodes. It utilizes the accumulated cross-episode feedback to compute a meta-gradient that updates the Agent's policy initialization $\theta^{(0)}$. By optimizing $\theta^{(0)}$, DeMP anticipates the Observer's learning trajectory, thereby mitigating the adaptation lag in RDPP.} \label{methodfig}

\end{figure*}

\section{Related Work}
\subsection{Deceptive planning}

Deception is a fundamental concept in adversarial scenarios, closely tied to privacy, security, and obfuscation. A general theory of deception~\cite{Whaley1982TowardAG,Bell2003TowardAT} defines simulation and dissimulation as two basic strategies. Building on this foundation, deception has been applied to path planning. In deceptive path planning (DPP), an action is considered deceptive if the real goal does not dominate the probability of other goals~\cite{masters_deceptive_2017}, and deception emerges as an optimal planning objective~\cite{huang2021dynamic,karabag2019optimal}.

Recent studies have introduced reinforcement learning (RL) frameworks for DPP, generating action sequences that keep the agent's true goal ambiguous among alternatives, such as the Ambiguity Model (AM)~\cite{liu_deceptive_2021} and the Deceptive Exploration Ambiguity Model (DEAM)~\cite{lewis_deceptive_2023}. These methods assume that the observer model is known to the agent, which then plans ambiguous paths across multiple goals. An extended goal recognition framework for strategic deception~\cite{masters_extended_2021} further investigates how factors such as perception and memory affect susceptibility to deception.
In addition, the motion-deception game~\cite{rostobaya2023eater} examines misleading opponents’ inference, deceptive planning under resource allocation~\cite{chen2024deceptive} addresses balancing deception and cost-efficiency, and multi-agent deception modeling with Theory of Mind (ToM)~\cite{sarkadi2019modelling} emphasizes simulating the observer’s mental state for effective deception.

Beyond classical ambiguity-based approaches, several extensions to deceptive path planning (DPP) have been proposed. Some works develop new methodological frameworks, such as reinforcement learning with graph neural networks to train general DPP policies~\cite{fatemi_deceptive_2024}, maximum-entropy formulations for deception under stochastic uncertainty~\cite{savas_deceptive_2022}, and mixed-integer programming for magnitude-based deception in single-goal settings~\cite{xu_single_2020}. Others address practical challenges by incorporating terrain costs in complex environments~\cite{lenhard_deceptive_2023}, designing domain-independent strategies based on landmarks, centroids, and minimum coverage states~\cite{price_domain_2023}, and handling adversarial cost signal manipulations through deceptive reinforcement learning~\cite{huang2019deceptive}.

While these studies broaden the modeling and applicability of DPP, they remain confined to one-shot settings and do not address repeated interactions with adaptive observers. While these studies provide diverse methods for deceptive path planning, they all focus on single-shot interactions. None address the challenge of repeated deceptive path planning with learnable observers, which is the focus of this work.

\subsection{Goal Recognition}

The goal recognition (GR) problem is directly relevant to deceptive planning because the observer in DPP is essentially a goal recognizer: GR methods specify how an observer maps observed behavior to a probability distribution over candidate goals. Classical GR approaches are often cost-based~\cite{masters_cost-based_2019, Ramrez2009PlanRA,ramirez2010probabilistic,sohrabi2016plan}: by assuming near-optimal behavior, cost-divergence or Bayesian plan-recognition techniques compute posterior probabilities over goals and thereby serve as natural observer models for some deceptive-planning works\cite{liu_deceptive_2021,masters_deceptive_2017}.

However, these classical methods rely on an optimality assumption and are typically formulated for single-episode inference. Extensions that relax optimality improve robustness to sub-optimal behavior~\cite{masters_goal_2019-1,zhi2020online}, and some research incorporates learned or hybrid models to improve scalability and robustness: neuro-symbolic and deep learning approaches handle noise and missing observations and offer fast inference for large candidate sets~\cite{amado2018goal, amado_robust_2023,chiari_goal_2022,chiari_fast_2024}.

These methods focus on inference accuracy, efficiency, or robustness to observation noise, but they do not consider scenarios where the recognizer needs to adapt across repeated adversarial interactions.

For these reasons, rather than relying on an off-the-shelf GR algorithm, we construct a learnable goal recognizer based on neural network as the observer in our experiments. This recognizer follows the standard GR setting and is designed to update online, making it a suitable component  for studying RDPP and validating methods.

\section{Preliminaries}
\textbf{Markov decision process (MDP)}~\cite{puterman2014markov}.
An MDP is defined by a tuple
$M = (\mathcal{S}, \mathcal{A}, \mathcal{P}, r, \gamma)$,
where $\mathcal{S}$ is the state space, $\mathcal{A}$ is the action space,
$\mathcal{P}(s_t, a_t, s_{t+1})$ is the transition function defining the probability of moving from state $s_t$ to $s_{t+1}$ given action $a_t$,
$r(s_t, a_t, s_{t+1})$ is the reward received for executing action $a_t$ in state $s_t$ and transitioning to $s_{t+1}$,
and $\gamma \in (0, 1)$ is the discount factor.
The objective is to learn a decision policy  $\pi : \mathcal{S} \rightarrow \mathcal{A}$, which maximizes the value function:
$
V_{\pi}(s) = \mathbb{E}\left[\sum_{t \in T} \gamma^t r(s_t, a_t, s_{t+1})\right],
$
where $T = \{0, 1, \dots, H-1\}$ denotes the set of discrete time steps within an episode of finite horizon $H$,

\textbf{Deceptive MDP}~\cite{ornik2018deception}.
A deceptive MDP is defined as
\begin{equation*}
M_{DPP} = (\mathcal{S}, \mathcal{A}, \mathcal{P}, \mathcal{R}, r, \mathcal{B}, \mathcal{L}, \gamma),
\label{MDPP}
\end{equation*}
where $\mathcal{S}, \mathcal{A}, \mathcal{P}, r$, and $\gamma$ are the same as in a regular MDP,
$\mathcal{R}$ is a set of candidate reward functions, including the true reward function $r$ and at least one deceptive reward function,
$\mathcal{B}$ is the observer's belief-set, and
$\mathcal{L}(s_t, a_t, s_{t+1}, b_t)$ is a belief-induced reward function, defining the reward for executing action $a_t$ in state $s_t$ and transitioning to $s_{t+1}$ while incorporating the impact of the observer's belief $b_t$ about the agent's true reward function.

The objective of a deceptive MDP is to maximise the belief-induced value function:
\begin{equation*}
V_{\pi}(s) = \mathbb{E}\left[\sum_{t \in T} \gamma^t \mathcal{L}(s_t, a_t, s_{t+1}, b_t)\right].
\end{equation*}

\textbf{Observer Model}. AM~\cite{liu_deceptive_2021} models observer beliefs via cost-based plan recognition, which uses $Q$-differences:
\begin{equation*}
\Delta_{r_i}(\vec{\sigma}) = \sum_{(s_t, a_t) \in \vec{\sigma}} \left( Q_{r_i}(s_t, a_t) - \max_{a'_t \in A} Q_{r_i}(s_t, a'_t) \right).
\end{equation*}
If behaviour is optimal for a reward function $r_i$, then $\Delta_{r_i}(\vec{\sigma}) = 0$. Otherwise, $\Delta_{r_i}(\vec{\sigma}) < 0$ and decreases with more sub-optimality. AM estimates probabilities for the candidate reward functions via a Boltzmann distribution:
\begin{equation*}
P(r_i \mid \vec{\sigma}) = \frac{\exp(\Delta_{r_i}(\vec{\sigma}))}{\sum_{r_j \in \mathcal{R}} \exp(\Delta_{r_j}(\vec{\sigma}))} \cdot P(r_i),
\end{equation*}
where $P(r_i)$ is an estimate of the prior probability that $r_i$ is the real reward function. A less optimal sequence leads to a lower probability. This is a proxy for the observer beliefs.

\section{Method}
In this section, we first formalized the repeated deceptive path planning task. To address this challenge of evolving opponents in repeated deceptive path planning, we propose a novel method DeMP, a two-level optimization framework combining episode-level adaptation for rapid policy adjustment after each episode with meta-level updates that leverage cross-episode feedback to accelerate adaptation in future episodes.

\subsection{Modeling Repeated Deceptive Path Planning}

\subsubsection{Problem Definition}

We define a Repeated Deceptive Path Planning problem as the tuple
\[
M_{RDPP} = \big(M_{DPP}, \Phi, K\big),
\]
where $M_{DPP}$ is a deceptive MDP,
$\Phi$ is the parameter space of the observer's learnable model, and $K$ denotes the number of repeated interactions.
RDPP extends $M_{DPP}$ by introducing repeated interactions with a learnable observer.

In RDPP the observer is not static: its predictive model is parameterized by $\phi \in \Phi$ and evolves across the $K$ episodes according to a learning dynamic. The agent  operates within the DPP structure while facing an adaptive observer whose parameters $\phi$ change over repeated interactions.

\subsubsection{Interaction Protocol}
Each interaction episode $k \in \{1, \dots, K\}$ proceeds as follows.
Within episode $k$, the agent interacts with the environment defined by $M_{DPP}$ according to its deceptive policy, generating a complete trajectory
\[
\tau^{(k)} = \{(s_t, a_t)\}_{t=0}^{T_k},
\]
where $T_k$ denotes the length of episode $k$.
The observer only receives a partial trajectory prefix
\[
\zeta^{(k)} = \tau^{(k)}_{0:\lfloor \alpha T_k \rfloor}, \quad \alpha \in (0,1],
\]
and produces a predictive distribution over candidate goals
\[
\mathcal{O}(\zeta^{(k)};\phi^{(k)}) = P(G \mid \zeta^{(k)}; \phi^{(k)}), \quad G \subseteq S,
\]
from which a prediction is obtained
\[
\hat{G}^{(k)} = \arg\max_G P(G \mid \zeta^{(k)}; \phi^{(k)}).
\]

After the episode, the agent is provided with $\mathcal{O}(\zeta^{(k)};\phi^{(k)})$ as additional feedback, while the observer receives the true goal $G^*$ and the complete trajectory $\tau^{(k)}$ for parameter adaptation. The observer then updates its parameters according to the update operator $\mathcal{U}$:
\[
\phi^{(k+1)} \leftarrow \mathcal{U}(\phi^{(k)}, \tau^{(k)}, G^*).
\]

The interaction protocol induces a two-sided adaptation process: the observer updates its predictive model between episodes based on post-episode feedback, while the agent needs to adapt its deceptive trajectories in response to counter the evolving observer.

\subsubsection{Observer's Learning Dynamics}

We model the observer as a learnable recognition model parameterized by $\phi \in \Phi$. Its role is to estimate the probability of the agent's true goal $G^*$ given the observed partial trajectory. Prior to the repeated deceptive interaction, the observer is initialized through an offline training phase on non-deceptive trajectories, providing a reasonable goal-recognition prior.

During repeated interactions, the observer adapts $\phi$ to improve predictive performance. Formally, the adaptation in episode $k$ is expressed as minimizing the negative log-likelihood loss:
\[
\mathcal{L}_{\mathcal{O}}(\phi) = - \log P(G^* \mid \tau^{(k)}; \phi).
\]

In practice,the update operator $\mathcal{U}$ may be realized through maximum-likelihood estimation, Bayesian updates, or gradient-based optimization (e.g., $\phi^{(k+1)} = \phi^{(k)} - \eta \nabla_\phi \mathcal{L}_{\mathcal{O}}(\phi^{(k)})$, where $\eta$ is the learning rate). This abstraction captures a broad class of adaptive observers while introducing controlled, modelable non-stationarity into the planning problem.

\subsubsection{Planning Agent's Objective}

The agent's objective in RDPP extends the standard DPP formulation to repeated interactions. Formally, given $K$ episodes, the cumulative value of a policy $\pi$ is
\[
V_\pi = \mathbb{E}\Biggl[\sum_{k=1}^{K}\sum_{t=0}^{T_k} \gamma^t \, \mathcal{L}(s_t^{(k)}, a_t^{(k)}, s_{t+1}^{(k)}, b_t^{(k)})\Biggr],
\]
where $\mathcal{L}$ denotes the belief-induced reward depending on the observer's latent belief $b_t^{(k)}$.

In practice, the agent cannot access the observer’s belief during execution. Instead, it only receives the observer’s predictive distribution $\mathcal{O}(\zeta^{(k)};\phi^{(k)})$ after each episode, which serves as the sole feedback for adapting its policy across episodes. Consequently, solving RDPP requires exploiting cross-episode feedback to improve long-term deception performance rather than optimizing deception within individual episodes in isolation.

\subsubsection{Challenges of RDPP}
The core challenge of RDPP lies in the non-stationarity induced by the continual evolution of the observer model. Concretely, the observer's learning dynamics $\mathcal{U}$ iteratively update its predictive model $\phi^{(k)}$ across episodes, which progressively increases the difficulty of deception for the agent. This non-stationarity affects the agent's decision-making process through the belief-induced reward $\mathcal{L}$, which couples the agent's actions with the observer's beliefs $b_t^{(k)}$ in order to suppress the probability that the observer assigns to the true goal $G^*$. However, in practice $\mathcal{L}$ is unobservable and typically requires manual design to align with adversarial objectives.

A naive approach to this problem is to incorporate the observer's predictions into the reward signal and update the agent's policy episode by episode. While this enables some degree of adaptation, it is fundamentally limited: since the agent only adapts after the observer has already improved, it inherently lags behind the observer's updates. This incremental adaptation lag accumulates over multiple interactions, ultimately compromising the agent's ability to sustain deception in repeated encounters.

\begin{figure*}[h]
\begin{adjustwidth}{-10pt}{-10pt}
		\centering

\subfigure[Deceptiveness across 400 episodes]{\includegraphics[width=0.27\linewidth]{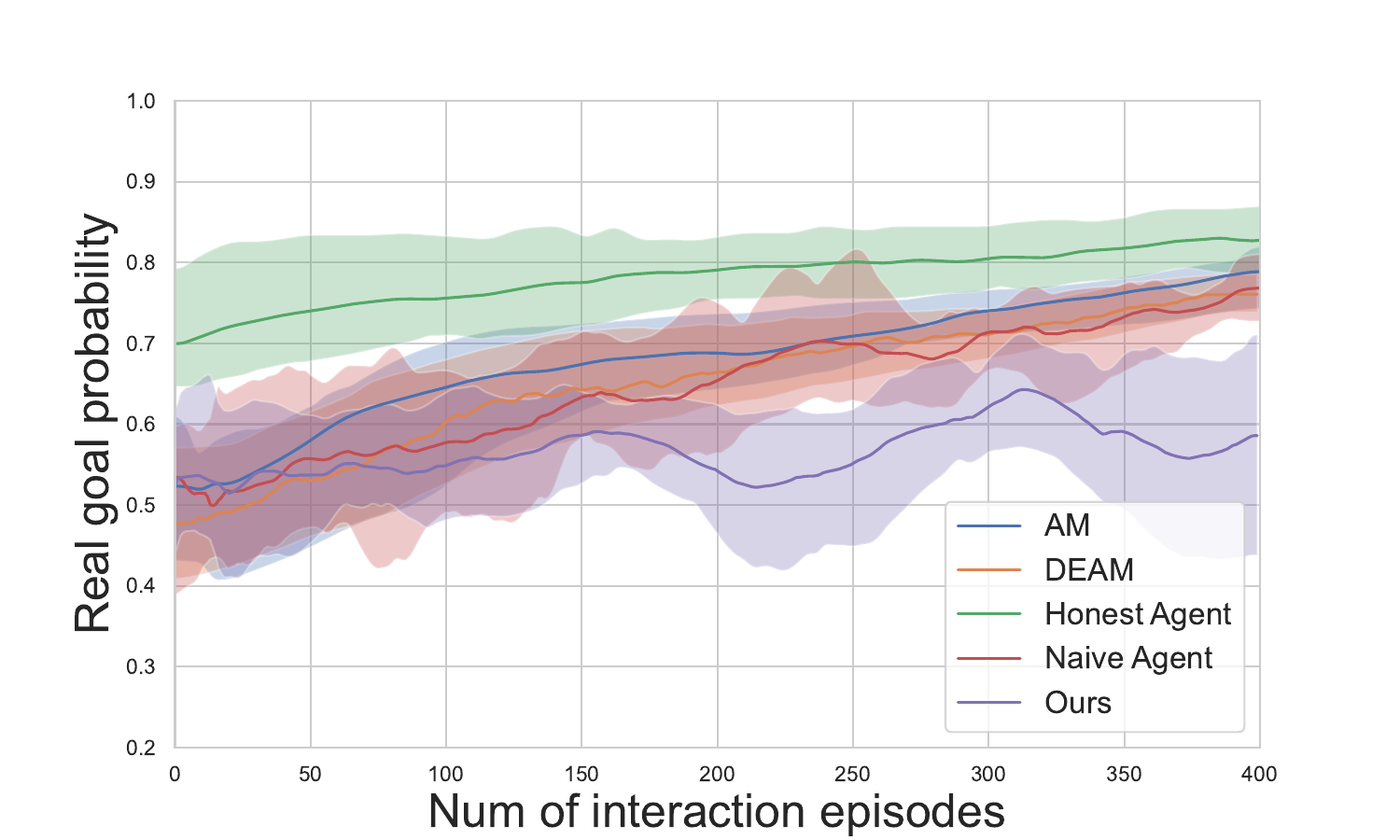}
}\hspace{-10pt}
\subfigure[Deceptiveness in last episode]{\includegraphics[width=0.27\linewidth]{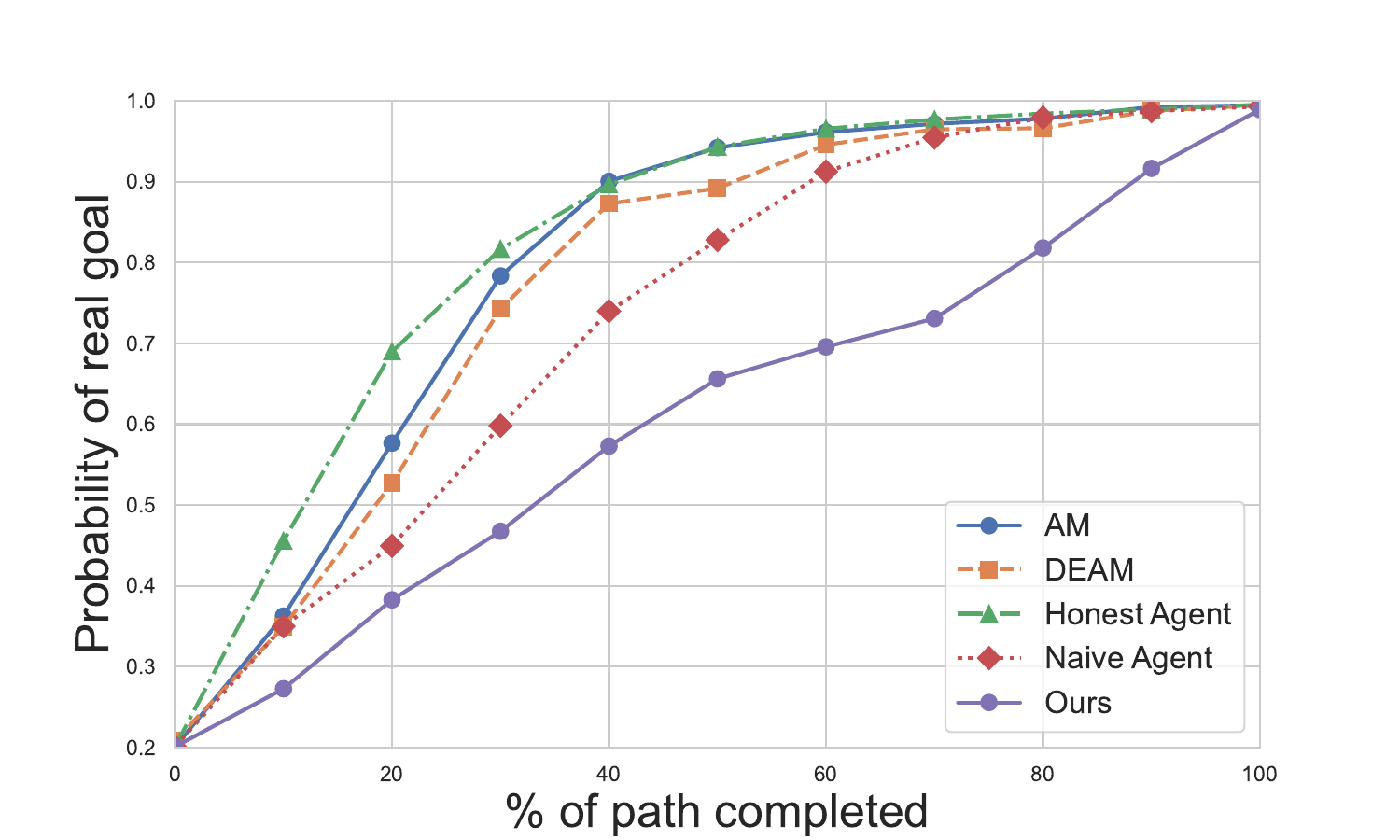}
}\hspace{-10pt}
\subfigure[Path costs]{\includegraphics[width=0.22\linewidth]{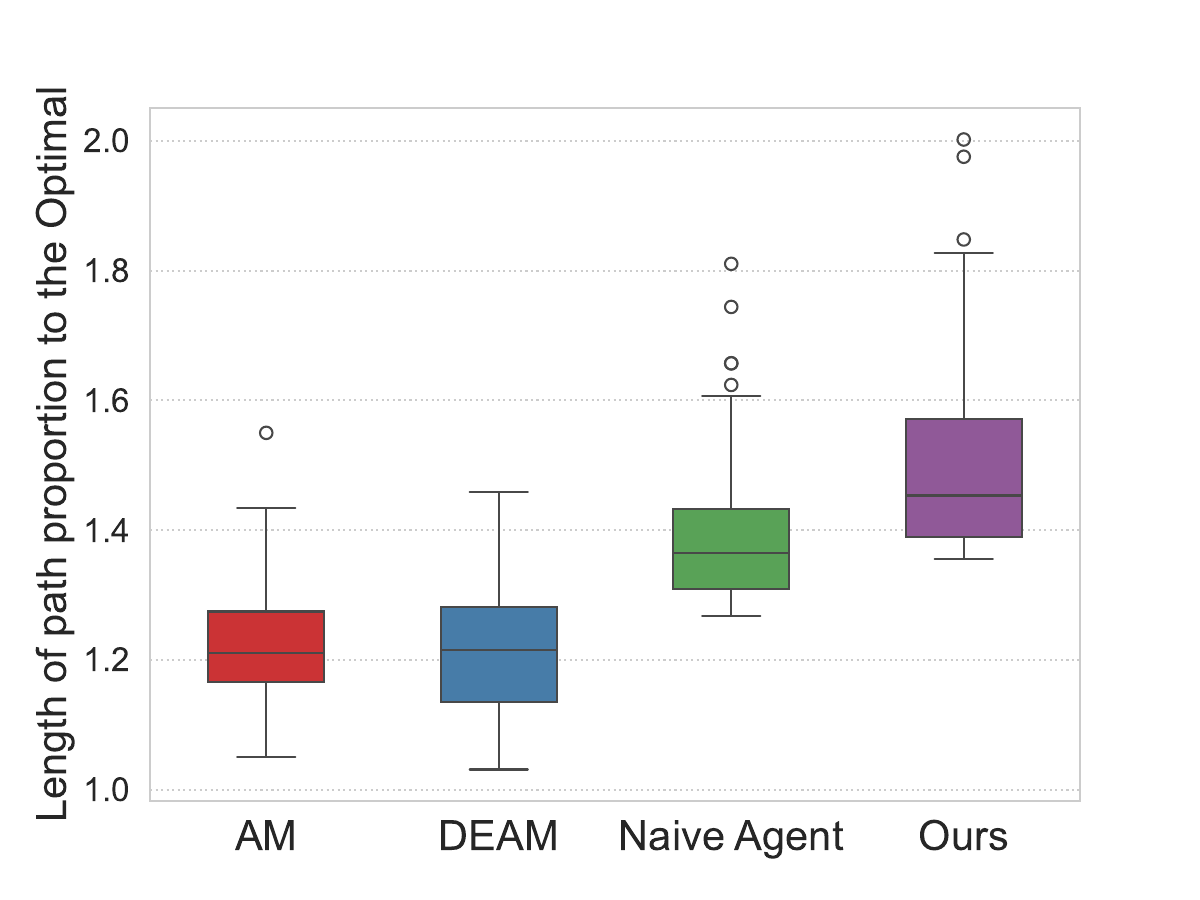}}
\subfigure[Steps-after-LDP]{\includegraphics[width=0.22\linewidth]{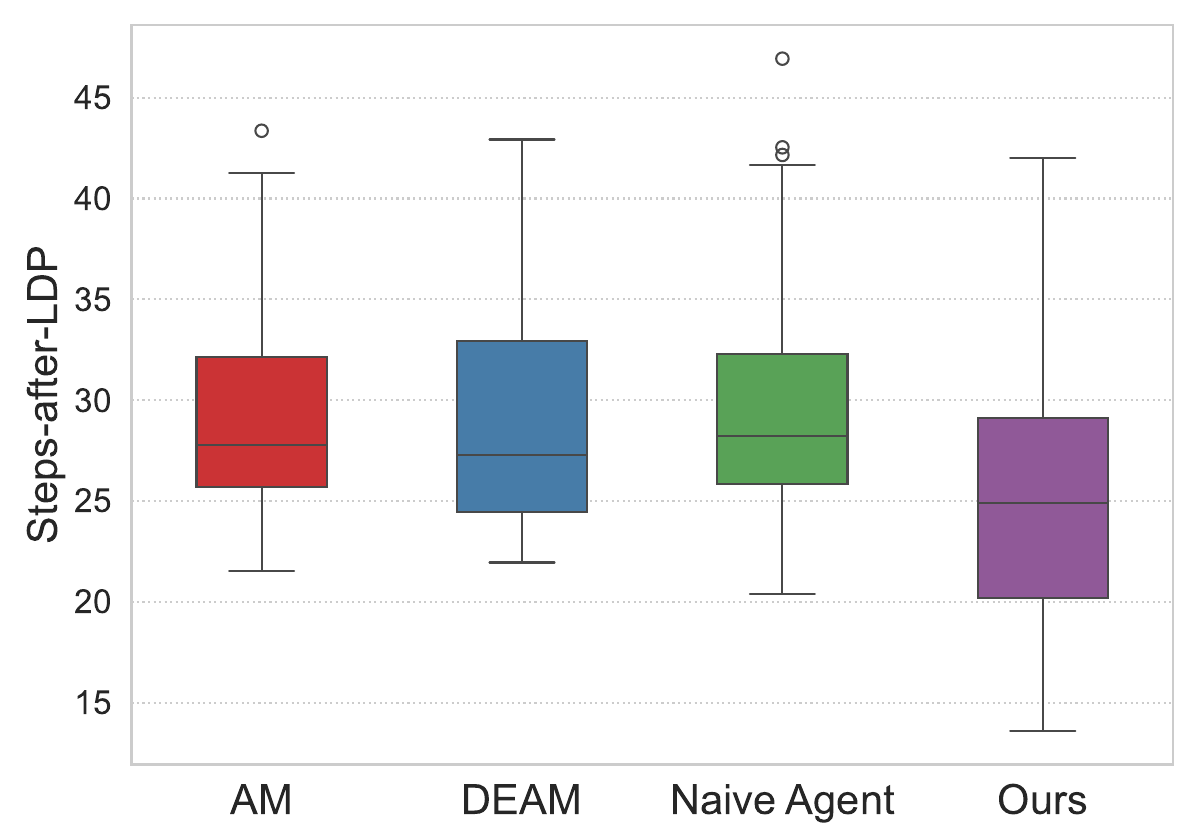}}\\[ -10pt]
\subfigure[Deceptiveness across 400 episodes]{\includegraphics[width=0.27\linewidth]{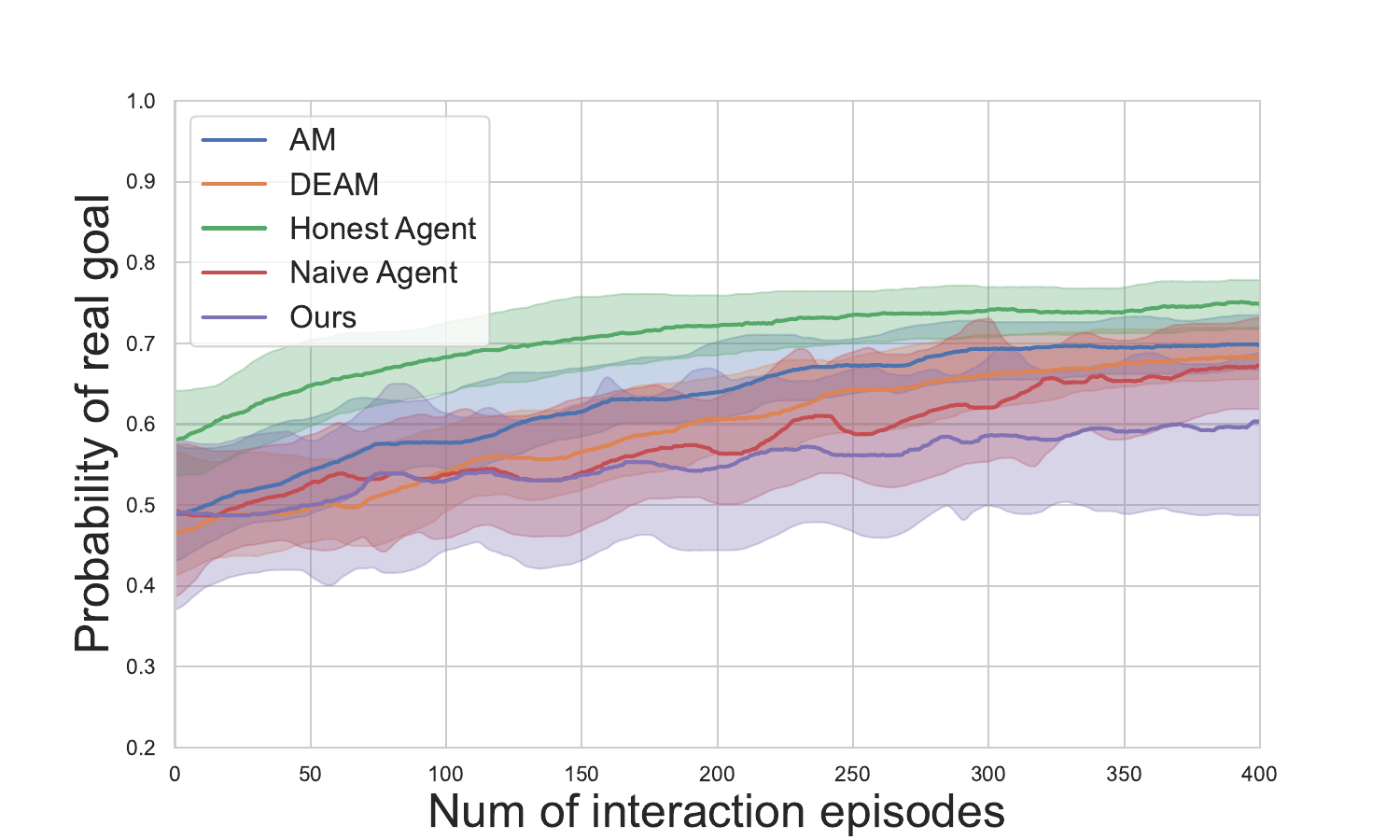}
}\hspace{-10pt}
\subfigure[Deceptiveness in last episode]{\includegraphics[width=0.27\linewidth]{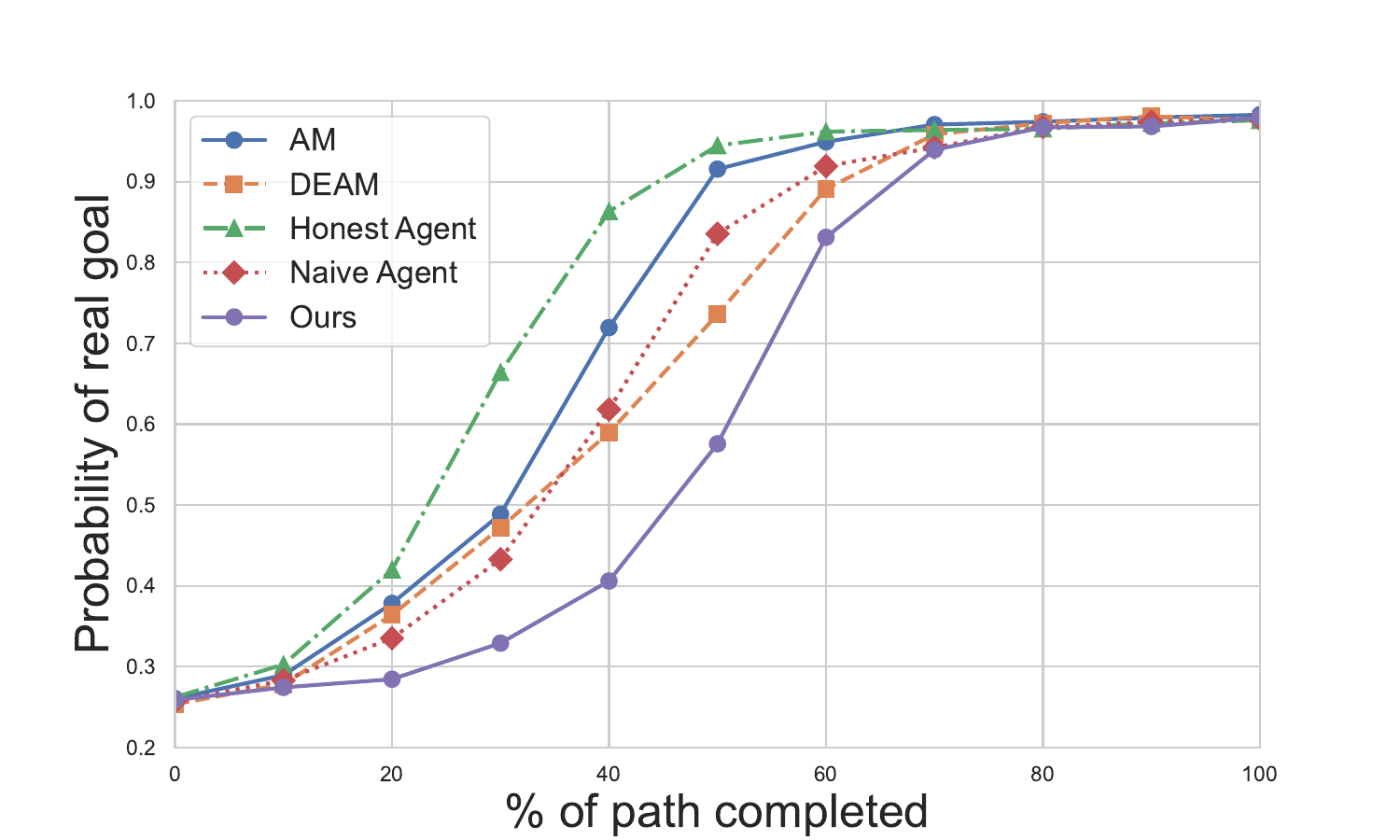}
}\hspace{-10pt}
\subfigure[Path costs]{\includegraphics[width=0.22\linewidth]{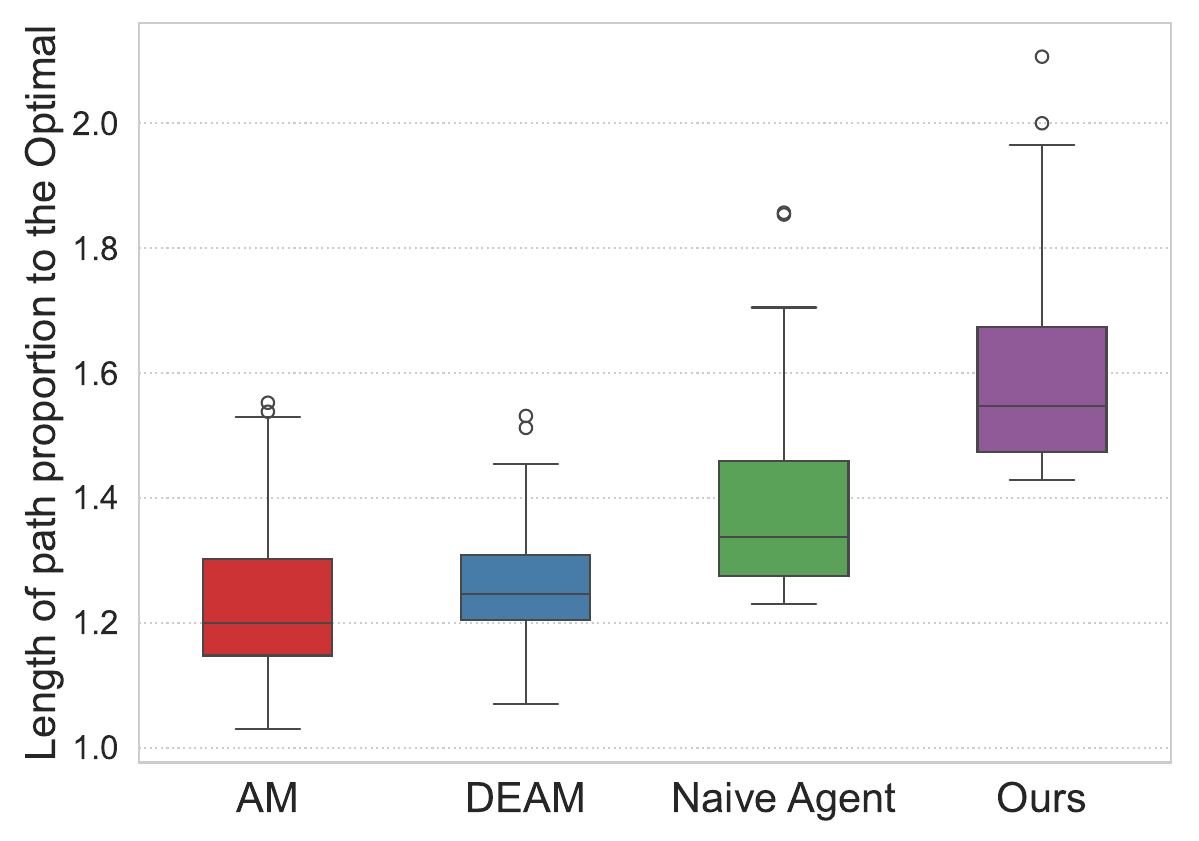}}
\subfigure[Steps-after-LDP]{\includegraphics[width=0.22\linewidth]{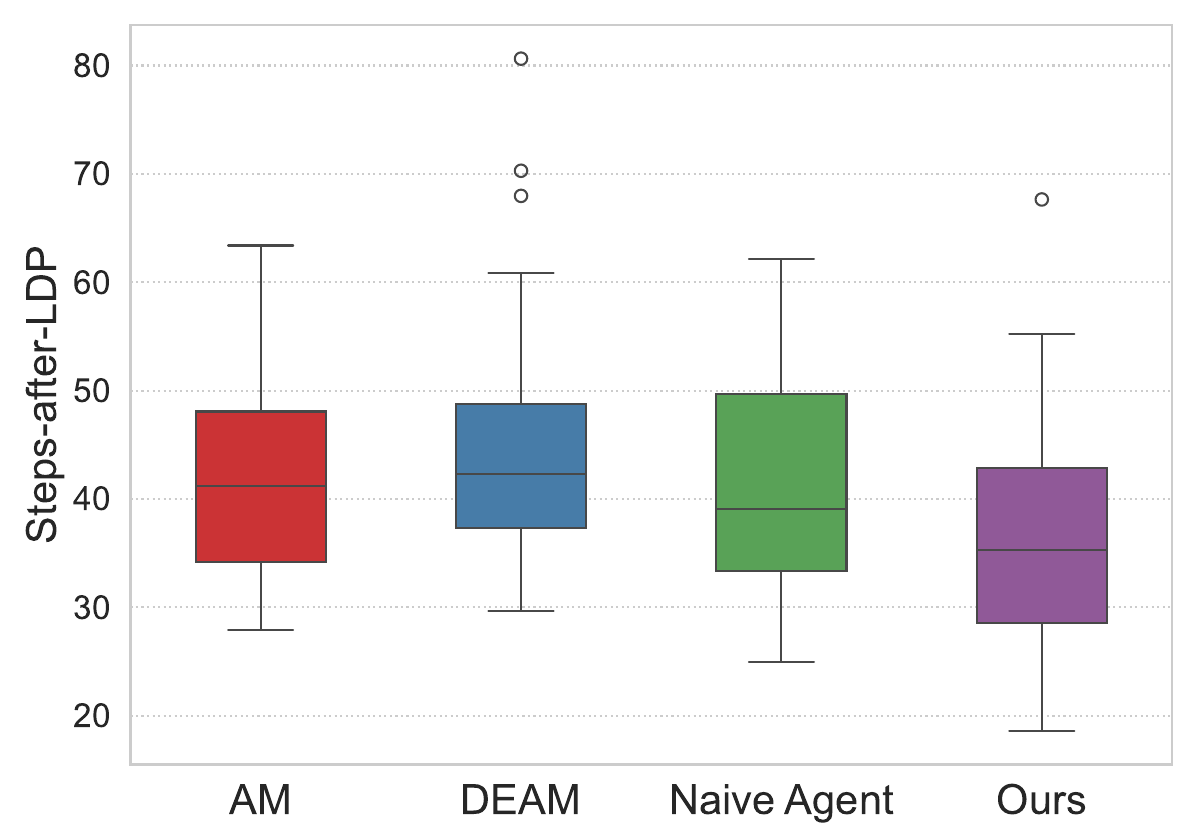}}\\[ -5pt]
\end{adjustwidth}
\caption{Deception performance and trajectory cost in repeated interactions.
The first row corresponds to the $49\times49$ grid environment, and the second row corresponds to the $100\times100$ grid environment.
(a,e) Deceptiveness across 400 episodes. DeMP consistently maintains the lowest probability assigned to the true goal.
(b,f) Deceptiveness in the last episode. After repeated interactions, baselines lose their deceptive effect early due to observer learning, while DeMP preserves low goal probability over most of the path.
(c,g) Path costs normalized by the optimal path length. DeMP incurs slightly higher cost due to continual exploration.
(d,h) Steps after the last deceptive point (LDP). DeMP keeps the observer uncertain for a longer portion of the trajectory.}

 \label{curvefig}
\vspace{-1em}
\end{figure*}

\subsection{Deceptive Meta Planning}

To overcome these challenges,  we propose \textbf{Deceptive Meta Planning (DeMP)}, a two-level optimization framework. As shown in Figure~\ref{methodfig}, the framework consists of an \textit{episode-level} adaptation, which adjusts the agent's policy after each episode based on the observer's updated predictions, and a \textit{meta-level} update that leverages cross-episode feedback to refine the agent's policy initialization for future episodes. Concretely, DeMP operates in a nested structure: between two consecutive meta-level updates, the agent engages in a sequence of $M$ episodes of interaction, within which an episode-level adaptation is applied after each episode. The aggregated feedback from these episodes then drives the subsequent meta-level update.
We next describe these two levels in detail.

\subsubsection{Episode-Level Adaptation}
Episode-level adaptation in DeMP corresponds to optimizing the parameters of the planning policy based on feedback from the observer.
This adaptation occurs after each episode of interaction, where the agent refines its policy to better maintain deception against the observer's updated model.

Concretely, within a sequence of $M$ consecutive episodes, the agent begins with initial parameters denoted as $\theta^{(0)}$ and performs an  episode-level adaptation after each episode.

At the very beginning of episode, $\theta^{(0)}$ is obtained from a pretrained deceptive planning agent
(e.g., a reproduction of an end-to-end AM~\cite{liu_deceptive_2021} model or another DPP agent trained on the standard deceptive MDP).
This initialization provides a reasonable starting point for online adaptation and in subsequent meta-updates, $\theta^{(0)}$ is updated to incorporate experience accumulated from previous sequences of episodes.

During the $k$-th episode ($k = 0, \dots, M-1$), the agent executes its policy $\pi_{\theta^{(k)}}$ in the environment, producing a trajectory
\[
\tau^{(k)} = (s_0, a_0, \dots, s_{T_k}).
\]
At the end of the episode, the agent receives the observer's predictive distribution over candidate goals $\mathcal{O}(\zeta^{(k)}; \phi^{(k)})$ as feedback for policy adaptation.

The episode-level objective is defined as a reinforcement learning loss
\[
\mathcal{L}_{\text{RL}}^{(k)}(\theta)
= - \mathbb{E}_{\tau^{(k)} \sim \pi_{\theta}} \big[ R^{(k)}(\tau^{(k)}) \big],
\]
where the total return
\[
R^{(k)}(\tau^{(k)})
= \sum_{t=0}^{T_k-1} r_{\text{env}}(s_t,a_t)
+ r_{\text{deceptive}}^{(k)}(G^*)
\]
combines the environment reward with a terminal deceptive reward
\[
r_{\text{deceptive}}^{(k)}(G^*)
= \big(1 - P(G^* \mid \zeta^{(k)}; \phi^{(k)})\big)\, r(G^*).
\]
This surrogate formulation penalizes the agent when the observer assigns high confidence to the true goal.

To further encourage deception, we add an observer-guided regularization term that minimizes the divergence between the observer's posterior and a target uniform confusion distribution $\mathrm{Unif}(G)$:
\begin{equation*}
\mathcal{L}^{(k)}(\theta)
= \mathcal{L}_{\text{RL}}^{(k)}(\theta)
+ \lambda \, \mathrm{KL}\!\left(\mathcal{O}(\zeta^{(k)};\phi^{(k)}) \,\big\|\, \mathrm{Unif}(G)\right).
\end{equation*}
The KL regularization  provides a tractable surrogate for maximizing belief entropy in standard belief-induced RDPP objectives.

After each episode, the agent updates its parameters via gradient descent:
\begin{equation}
\theta^{(k+1)} = \theta^{(k)} - \alpha \nabla_{\theta^{(k)}} \mathcal{L}^{(k)}(\theta^{(k)}),
\label{eq:episode_update}
\end{equation}
where $\alpha$ is the episode-level learning rate.

After $M$ consecutive episodes, the agent obtains the adapted parameter $\theta^{(M)}$
  by adapting its policy after each episode of interaction to counter the observer's updated recognition model.

\subsubsection{Meta-Level Update}

After $M$ episode-level updates, we define the meta-objective based on the final adapted parameter:
\begin{equation}
\mathcal{L}^{(\text{meta})}(\theta^{(0)})
= \mathcal{L}^{(M)}\big(\theta^{(M)}(\theta^{(0)})\big),
\label{eq:meta_loss}
\end{equation}
where $\theta^{(M)}(\theta^{(0)})$ denotes the parameter obtained after performing $M$ consecutive episode-level updates starting from initialization $\theta^{(0)}$. In other words, $\theta^{(M)}(\theta^{(0)})$ is treated as a function mapping the initial parameter to the result of $M$ gradient steps. The gradient of the meta-objective with respect to the initialization $\theta^{(0)}$ can be computed via the chain rule:
\begin{equation}
\nabla_{\theta^{(0)}} \mathcal{L}^{(\text{meta})}
= \nabla_{\theta^{(M)}} \mathcal{L}^{(M)} \cdot \frac{\partial \theta^{(M)}}{\partial \theta^{(0)}}.
\label{eq:meta_grad}
\end{equation}

Finally, the meta-update is performed as:
\begin{equation}
\theta^{(0)} \leftarrow \theta^{(0)} - \beta \, \nabla_{\theta^{(0)}}\mathcal{L}^{(\text{meta})}(\theta^{(0)}),
\label{eq:meta_update}
\end{equation}
where $\beta$ is the meta-level learning rate.  In the meta-level update, DeMP integrates long-term feedback across multiple episodes using higher-order gradient information. This allows DeMP to effectively capture patterns in how the observer updates its predictions over time and to refine the initialization such that future adaptation (in the next $M$-episode block) can start from a more effective parameter point.

This meta-level mechanism is structurally related to meta-learning approaches such as MAML~\cite{finn2017model}, in that both optimize an initialization for rapid adaptation. Unlike MAML, however, where adaptation occurs across a distribution of independent tasks, DeMP adapts within a single deceptive planning task against an evolving observer, making the source of variation endogenous to repeated interactions.

The overall optimization alternates between episode-level adaptation and meta-level updates, as summarized in Algorithm~\ref{alg:DeMP}. By refining the policy initialization based on cross-episode feedback, DeMP enables faster adaptation in subsequent interaction blocks and mitigates the accumulation of adaptation lag in repeated deceptive planning.

\begin{algorithm}[h!]
\caption{Deceptive Meta Planning (DeMP)}
\label{alg:DeMP}
\begin{algorithmic}[1]
\State Initialize agent parameters $\theta^{(0)}$ from pretrained DPP agent
\While{online interaction continues}
    \For{$k = 0$ to $M-1$}
        \State Start a new episode
        \State Execute policy $\pi_{\theta^{(k)}}$ to collect trajectory $\tau^{(k)}$
        \State Obtain observer's prediction $\mathcal{O}(\zeta^{(k)}; \phi^{(k)})$
        \State Compute episode loss $\mathcal{L}^{(k)}(\theta^{(k)})$
        \State Update policy parameters $\theta^{(k+1)}$ according to episode-level adaption
    \EndFor
    \State Compute meta loss $\mathcal{L}^{(\text{meta})}(\theta^{(0)})$
    \State Update initialization $\theta^{(0)}$ according to meta-level update
\EndWhile
\end{algorithmic}
\end{algorithm}

\subsection{Theoretical Analysis}
\label{sec:theoretical_analysis}

A central challenge in RDPP is that the agent must optimize against an observer whose belief state is latent and whose parameters evolve over time. We provide a theoretical justification for the validity of DeMP, with formal proofs deferred to \textbf{Appendix A}.

\paragraph{Validity of the Surrogate Objective.}
RDPP theoretically requires maximizing a belief-induced reward defined on the observer’s latent posterior $b(G^* \mid \tau)$. In contrast, DeMP optimizes a surrogate objective constructed from the observer’s explicit prediction $P(G^* \mid \tau; \phi)$ together with an entropy regularization term.
This substitution is justified under the assumption of a \emph{rational observer} (see Assumption~1 in Appendix~A.1), which requires that the observer’s predictive model preserves the uncertainty ordering of its internal belief.

\begin{theorem}[Surrogate Consistency, Informal]
Under rational observer assumptions, minimizing the DeMP surrogate loss is optimization-consistent with maximizing the RDPP belief-induced reward.
\end{theorem}

As shown in Theorem~1 (Appendix~A.2), suppressing the predicted probability $P(G^*)$ serves as a monotonic proxy for reducing the true belief assigned to the goal, while minimizing the KL divergence to a uniform distribution provides a tractable surrogate for maximizing belief entropy. Together, these terms align the DeMP optimization landscape with the fundamental RDPP objective.

\paragraph{Mitigation of Adaptation Lag.}
In RDPP, purely reactive optimization suffers from \emph{adaptation lag}, as the policy is optimized for the current observer $\phi^{(k)}$ but evaluated after the observer updates to $\phi^{(k+1)}$. DeMP addresses this issue through meta-level optimization.

\begin{theorem}[Anticipation Mechanism, Informal]
The meta-gradient update in DeMP incorporates a second-order correction that reduces sensitivity to the observer’s future learning dynamics.
\end{theorem}

As formalized in Theorem~2 (Appendix~A.3), the meta-gradient with respect to the policy initialization includes a term that explicitly depends on the observer’s learning gradient. This term captures the anticipation of the observer’s update, enabling DeMP to learn policy initializations that are robust to belief evolution and thereby mitigating adaptation lag compared to purely reactive baselines.

\section{Experiment}

We evaluate the proposed Deceptive Meta Planning (DeMP) framework in repeated deceptive path planning (RDPP) tasks, aiming to quantify both its sustained deception performance and trajectory efficiency compared to existing methods.

\paragraph{DeMP Implementation.}

DeMP is implemented using the Soft Actor-Critic (SAC) algorithm~\cite{haarnoja2018soft}.
Both the policy and value networks are two-layer multilayer perceptrons (MLPs).
The agent first performs imitation learning on trajectories generated by AM~\cite{liu_deceptive_2021} to initialize a deceptive planning policy.
During online interaction, DeMP performs two-level optimization: an episode-level update after each interaction and a meta-level update every $M=2$ episodes.
The learning rates are set to $\alpha = 0.001$ for episode-level adaptation and $\beta = 0.0001$ for meta-level updates.
In each episode, the observer only receives a randomly proportion sampled from $[40\%, 60\%]$ prefix of the agent's trajectory before producing its goal recognition.
Each evaluation sequence includes 400 episodes of online interaction, and all experiments are independently repeated 100 times with different random seeds to ensure the robustness of the results.

\paragraph{Baselines.}
We compare DeMP with the following methods:
(1)~an \textbf{Honest Agent} that plans directly toward the true goal;
(2)~the \textbf{AM}~\cite{liu_deceptive_2021}  method;
(3)~the \textbf{DEAM}~\cite{lewis_deceptive_2023} method; and
(4)~a \textbf{Naïve Approach}, which adopts the same SAC framework and initialization as DeMP but directly incorporates the observer's predicted goal probability into the reward and performs only per-episode policy updates.

\paragraph{Experimental Settings.}
We evaluate DeMP under two standard deceptive planning scenarios adapted to the RDPP setting.
In the \textbf{Standard Path Deception Setting}, experiments are conducted on $49\times49$ and $100\times100$ grid maps with five candidate goals and large obstacle structures, which emphasize long-horizon planning and scalability under repeated interactions.
In the \textbf{Pirate Deception Setting}~\cite{nichols2022adversarial}, experiments are conducted on $49\times49$ grids with five candidate goals and high-density random obstacles, where an adversarial pirate actively pursues the agent based on the observer’s predictions.
Unless otherwise stated, reward structures follow standard grid-based navigation conventions.

\paragraph{Observer Model and Metrics.}
The observer is a two-layer LSTM model that outputs a probability distribution over candidate goals given the observed partial trajectory.
Before evaluation, it is pre-trained on non-deceptive trajectories collected from the same maps to develop goal-recognition ability, and then fine-tuned online via supervised learning during interaction. Additional details and empirical analysis of the observer pretraining process are provided in Appendix B.2.

We evaluate models using (1) the observer's predicted probability of the true goal (deceptiveness), (2) the trajectory length ratio relative to the optimal path (path cost), and (3) steps after last deceptive point (LDP).

\begin{figure}[t]
		\centering
\subfigure{\includegraphics[width=0.48\linewidth]{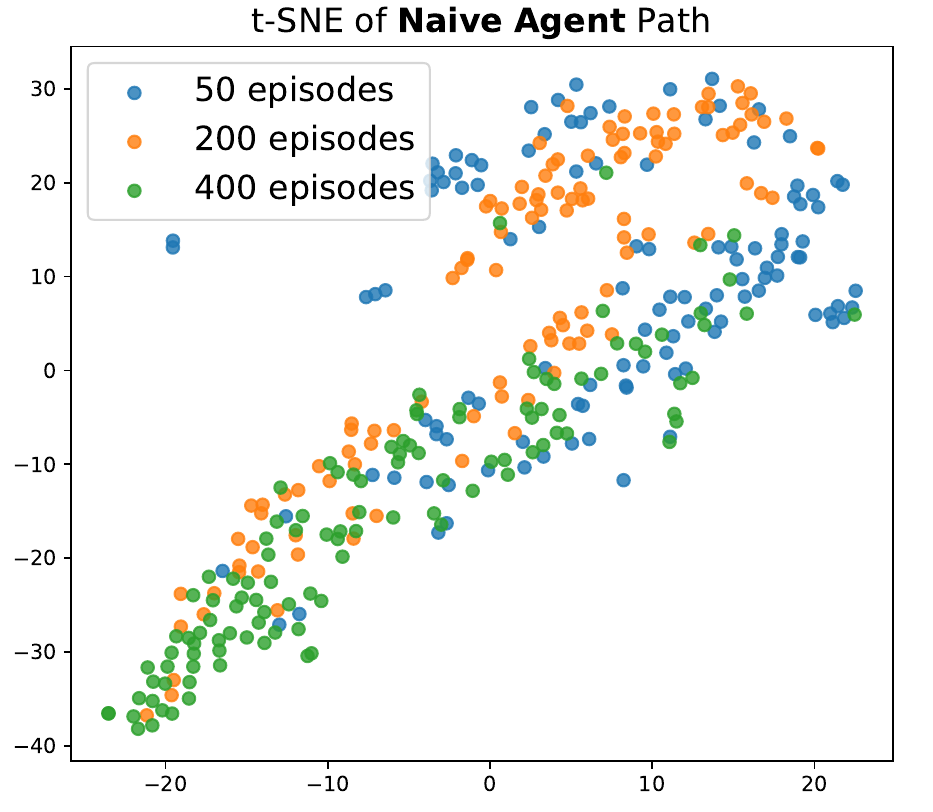}}
\hspace{-5pt}
\subfigure{\includegraphics[width=0.495\linewidth]{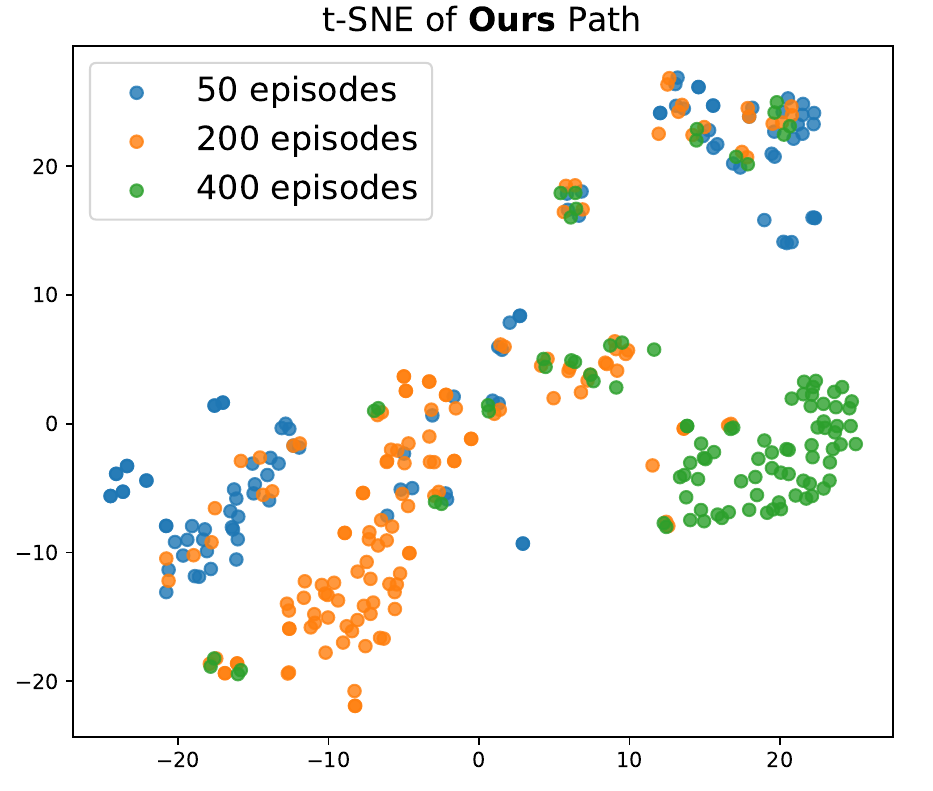}}
\vspace{-5pt}
		\caption{ t-SNE Projection of Path Features. The figure compares the distribution of generated path features over 400 episodes. The Naive Agent (Left) shows a highly concentrated distribution, indicating a rigid strategy. In contrast, DeMP (Right) exhibits significant path variability, with trajectories dispersing into new clusters, demonstrating continuous policy adaptation over extended interactions.
        } \label{tsnefig}
        \vspace{-1.0em}
\end{figure}

\subsection{Results of Standard Path Deception Setting}

We first evaluate all methods under the Standard Path Deception Setting with five candidate goals and large obstacles.
Figure~\ref{curvefig}a and \ref{curvefig}e reports the observer’s predicted probability of the true goal over 400 repeated interactions.
DeMP consistently maintains the lowest true-goal probability throughout the entire interaction horizon, while all baseline methods exhibit a steady degradation in deceptive performance as the observer adapts.

Initially, AM achieves relatively strong deception because the observer is pre-trained on optimal trajectories and has limited exposure to deceptive behaviors.
However, AM relies on a largely fixed path structure, which quickly becomes exploitable as the observer accumulates experience across interactions.
Both the Naïve Agent and DeMP start from the same honest initialization, but differ fundamentally in their adaptation mechanisms: the Naïve Agent updates its policy only at the episode level, whereas DeMP additionally leverages meta-level updates across episodes.
This meta-level adaptation allows DeMP to anticipate how the observer evolves and to maintain deceptive effectiveness over long-term interactions.

Figure~\ref{curvefig}b and \ref{curvefig}f further illustrates the observer’s belief evolution along the final episode trajectories.
For baseline methods, the observer rapidly identifies the true goal early in the trajectory, reflecting the formation of stable associations between repeated trajectory prefixes and goals.
In contrast, DeMP suppresses early goal recognition by continually altering its deceptive strategies, preventing the observer from converging to a reliable inference pattern. Figure~\ref{curvefig}c and \ref{curvefig}g shows that this sustained deception comes at the cost of slightly longer paths, representing a natural trade-off between efficiency and long-term robustness. Moreover, DeMP maintains deception over a larger portion of the trajectory, as reflected by the steps-after-LDP metric in Figure~\ref{curvefig}d and \ref{curvefig}h.

Figure~\ref{heatfig} visualizes the trajectory evolution of DeMP. The leftmost panel shows the static path distribution of the AM method, while the remaining heatmaps illustrate DeMP's trajectory distributions at different interaction stages. The bright regions indicate frequently visited states, and the dashed line denotes the final trajectory. Over time, DeMP's paths evolve from near-optimal routes to increasingly diverse patterns, demonstrating dynamic path adaptation and repeated deception against the learnable observer.

Finally, Figure~\ref{tsnefig} presents t-SNE projections of the path features. Each point represents a trajectory encoded by uniformly sampled 2D waypoints and projected into a two-dimensional space using t-SNE, where distances reflect trajectory similarity. The Naïve Agent's trajectories cluster tightly, indicating a static or rigid strategy with limited adaptation. In contrast, DeMP's trajectories form distinct clusters that expand and separate as interactions progress, reflecting the generation of new and diverse trajectories, consistent with the expected behavioral diversity induced by meta-level adaptation.

\subsection{Results of Pirate Deception Experiment}

To further evaluate the robustness of deceptive strategies under active pursuit, we extend our analysis to the Pirate Deception Scenario~\cite{nichols2022adversarial}.
 In this scenario, while the agent moves toward its goal, pirates are also active in the environment, attempting to infer the agent's true target and intercept it. The pirates rely on the candidate goal with the highest predicted probability from the observer as their pursuit target and move toward it following a rule-based shortest path. The pirates share the same discrete action space as the agent, with both parties allowed one move per step, and their initial positions are randomly assigned at the beginning of each episode.

\begin{figure}[t]
\begin{adjustwidth}{-5pt}{-5pt}
    \centering
    \subfigure[]{
        \includegraphics[width=0.46\linewidth]{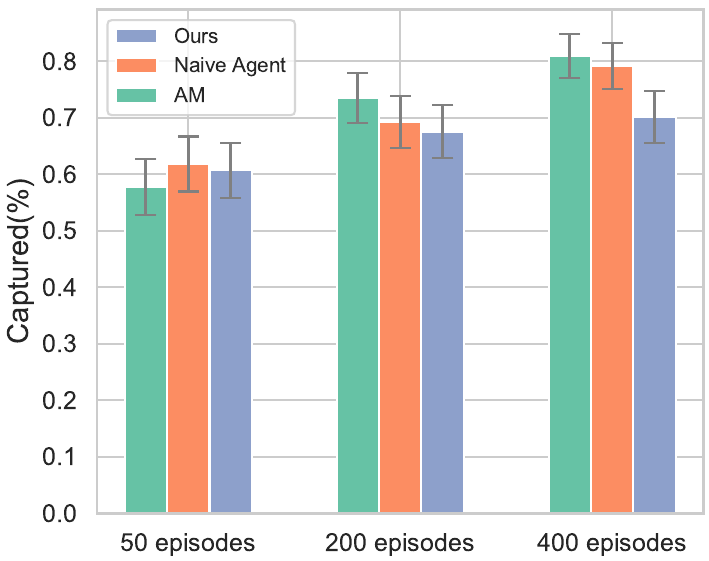}
        \label{fig:pirate_stat}
    }
    \subfigure[]{
        \includegraphics[width=0.46\linewidth]{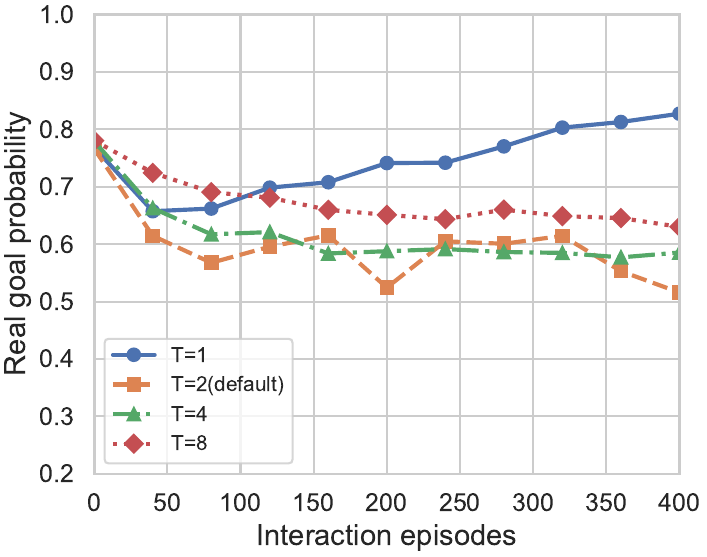}
        \label{fig:M_ablation}
    }
\end{adjustwidth}
\vspace{-10pt}
    \caption{Analysis of DeMP under repeated deceptive planning.
    (a) DeMP consistently achieves the lowest capture rate in the Pirate Deception Scenario as interactions increase.
    (b) Sensitivity analysis of the meta-episode parameter $M$, illustrating the trade-off between adaptation frequency and long-term deception performance.}
    \label{fig:pirate_and_M}
    \vspace{-1em}
\end{figure}

Agents and observers after 50, 200, and 400 interactions from the previous experiment were used in this evaluation. For each configuration, we conducted 100 random trials under fixed parameters for both sides.
Figure~\ref{fig:pirate_and_M}(a) reports the agent capture probability as interactions increase.
All methods exhibit rising capture rates as the observer becomes more accurate over time.
However, the degradation is substantially slower for DeMP, which consistently achieves the lowest capture probability across all interaction stages.
In contrast, AM and the Naïve Agent suffer rapid performance deterioration, indicating that their deceptive strategies become increasingly exploitable under repeated adversarial observation.
These results demonstrate that meta-level adaptation enables DeMP to maintain robust deception even under active pursuit.

\subsection{Analysis of Meta Episode Parameter $M$}
The meta episode parameter $M$ controls the number of consecutive DPP episodes considered before a meta-level update is applied. A larger $M$ allows the model to incorporate more historical interaction information, potentially enhancing adaptability to the evolving observer. However, increasing $M$ also reduces the frequency of meta-level updates, which may introduce a delay in adjusting the episode-level policy initialization.

To investigate the impact of $M$ on performance, we evaluated DeMP under different values of $M$. As shown in Figure~\ref{fig:pirate_and_M}(b), when $M=1$, the meta-level update degenerates to a per-episode update, causing the model's performance to approximate that of the Naïve Agent. In this case, deception performance decreases over repeated interactions since only single-episode information is considered.

For larger values of $M$, the model consistently maintains or improves deception performance. In particular, intermediate values of $M$ achieve a balance between incorporating sufficient historical information and maintaining timely adaptation, resulting in relatively stable deception across interactions. However, excessively large $M$ values lead to less frequent meta-level updates, causing delayed adaptation and slight deterioration in performance. Additionally, larger $M$ increases computational cost due to the need to store and backpropagate through longer episode sequences.

\section{Conclusion}

In this work, we introduced the task of repeated deceptive path planning (RDPP) and proposed Deceptive Meta Planning (DeMP), a novel two-level optimization framework. DeMP performs episode-level policy adaptation guided by an observer's feedback and meta-level updates to refine the initial policy, enabling sustained deception across repeated interactions. Theoretical analysis confirms that DeMP aligns with belief-induced RDPP objectives and proactively adapts to observer dynamics. Extensive experiments across grid and continuous domains demonstrate that DeMP significantly outperforms baselines, sustaining deception against evolving observers.

Despite these promising results, our study has certain limitations. The experiments do not capture the full complexity and dynamics of real-world planning problems. The effectiveness of DeMP in more complex or high-dimensional domains remains to be explored.

For future work, we aim to extend DeMP to more realistic and dynamic planning scenarios, consider multiple interacting agents, and investigate the integration of richer observer models. These directions could further enhance the adaptability and robustness of deceptive planning in practical applications.

\begin{acks}
This work was supported by the National Science and Technology Major Project under Grant No. 2022ZD0116403, and in part by the Beijing Natural Science Foundation under Grant No. 4264131.
\end{acks}
\bibliographystyle{ACM-Reference-Format}
\bibliography{sample}

\clearpage


\section*{Supplementary material}

# Repeated Deceptive Path Planning against Learnable Observer
# Appendix

## A Theoretical Analysis

In this appendix, we provide the theoretical justification for the Deceptive Meta Planning (DeMP) framework. We first formalize the assumptions regarding the observer's behavior. We then prove that the surrogate objective optimized by DeMP aligns with the fundamental objective of Repeated Deceptive Path Planning (RDPP). Finally, we analyze how the meta-level update mitigates the adaptation lag inherent in the interaction protocol.

### A.1 Assumptions

To bridge the gap between the observer's latent belief $b(G \mid \zeta)$ (which dictates the theoretical RDPP reward) and its output prediction $\mathcal{O}(\zeta; \phi) = P(G \mid \zeta; \phi)$ (observed by the agent), we introduce the following assumptions.

**Assumption 1** (Rational Observer and Order Preservation)**.** *The observer updates its parameters $\phi$ by minimizing a proper scoring rule (specifically, the negative log-likelihood $\mathcal{L}_{\mathcal{O}}$ defined in the main text). We assume the observer is sufficiently rational such that for any two trajectories $\tau_1, \tau_2$, if the observer assigns a lower probability to the true goal for $\tau_1$ compared to $\tau_2$, the observer's internal belief state follows the same order:*

$$P(G^* \mid \zeta_1; \phi) < P(G^* \mid \zeta_2; \phi) \implies b(G^* \mid \zeta_1) < b(G^* \mid \zeta_2).$$

*Remark on Assumption 1.* In the RDPP setting considered in this paper, the observer is initialized through an offline pre-training phase before any deceptive interaction takes place, and is subsequently adapted online during repeated episodes. This design provides a stable goal-recognition prior and empirically supports the order-preservation assumption used in the analysis.

**Assumption 2** (Observer Learning Regularity)**.** *The observer's update operator $\mathcal{U}$ is differentiable and Lipschitz continuous with respect to the input trajectory. That is, a small perturbation in the agent's trajectory $\tau^{(k)}$ results in a bounded change in the observer's parameters $\phi^{(k+1)}$. This allows us to approximate the observer's parameters as locally stationary within the optimization of a single episode-level policy update.*

### A.2 Proof of Objective Alignment

The fundamental goal of RDPP is to maximize a belief-induced return that depends on the observer's evolving belief state. As defined in the main text, this return is expressed in a general per-step form that captures both task completion and deception through the belief $b_t^{(k)}$.

Under standard RDPP assumptions that deception is primarily determined by the observer's terminal belief and that belief updates along a trajectory are monotonic, we consider an equivalent instantiation of this objective that makes the dependence on the observer's belief explicit at the episode level for the purpose of theoretical analysis.

In particular, a canonical belief-induced objective for episode $k$ can be written as

$$J_{RDPP}^{(k)} = \mathbb{E}_{\tau \sim \pi} \left[ 1 - b(G^* \mid \zeta^{(k)}) + \lambda H(b(\cdot \mid \zeta^{(k)})) \right],$$

where the first term penalizes high belief assigned to the true goal $G^*$, and the entropy term $H(\cdot)$ encourages overall uncertainty in the observer's belief. This formulation is consistent with the general RDPP objective in the main text and serves as a convenient instantiation for analyzing objective alignment.

DeMP optimizes the following loss function (derived from the rewards defined in Eq. (6) and (8) of the main text):

$$\mathcal{L}^{(k)}(\theta) = \mathbb{E}_{\tau \sim \pi_\theta} \left[ -(1 - P(G^* \mid \zeta^{(k)}; \phi^{(k)}))r(G^*) + \lambda \mathrm{KL}(\mathcal{O} \| \mathrm{Unif}) \right].$$

**Theorem 1** (Surrogate Objective Consistency)**.** *Under Assumption 1, minimizing the DeMP loss $\mathcal{L}^{(k)}(\theta)$ is optimization-consistent with maximizing the RDPP objective $J_{RDPP}^{(k)}$.*

*Proof.* We analyze the two components of the DeMP loss function separately.

**1. Terminal Deceptive Reward:** The agent maximizes the terminal reward $r_{deceptive}^{(k)} \propto (1 - P(G^* \mid \zeta^{(k)}; \phi^{(k)}))$. Minimizing the associated loss is equivalent to minimizing the observer's predicted probability of the true goal $P(G^* \mid \zeta^{(k)}; \phi^{(k)})$. By Assumption 1 (Order Preservation), any policy update $\Delta\theta$ that reduces $P(G^* \mid \cdot)$ also reduces the true latent belief $b(G^* \mid \cdot)$. Thus, optimizing this surrogate directly improves the first term of $J_{RDPP}^{(k)}$.

**2. KL Regularization:** The regularization term minimizes $\mathrm{KL}(P(\cdot \mid \zeta^{(k)}; \phi^{(k)}) \| \mathrm{Unif})$. Recall the definition of KL divergence between a distribution $P$ and a uniform distribution $U$ over $|G|$ goals:

$$\begin{aligned} \mathrm{KL}(P \| U) &= \sum_{g \in G} P(g) \log \frac{P(g)}{1/|G|} \\ &= \sum_{g \in G} P(g) \log P(g) - \sum_{g \in G} P(g) \log(1/|G|) \\ &= -H(P) + \log |G|. \end{aligned}$$

Since $\log |G|$ is constant, minimizing the KL divergence is mathematically equivalent to **maximizing the entropy** $H(P)$ of the observer's predictive distribution. Since $P(\cdot \mid \zeta)$ is a rational estimate of the latent belief $b(\cdot \mid \zeta)$ (Assumption 1), maximizing the entropy of the predictive distribution prevents the observer from becoming overconfident in any specific outcome (including decoys). This serves as a tractable surrogate for maintaining high entropy in the true belief $H(b)$.

**Conclusion:** Combining both terms, minimizing $\mathcal{L}^{(k)}(\theta)$ drives the policy to generate trajectories that both suppress the probability of the true goal and increase the overall uncertainty (entropy) of the observer. Thus, the DeMP objective aligns with $J_{RDPP}^{(k)}$. □

## A.3 Analysis of Adaptation Lag Mitigation

In RDPP, a core challenge is the non-stationarity of the observer. A naive agent optimizes $\theta^{(k+1)}$ to minimize loss against the *current* observer $\phi^{(k)}$, but it is evaluated against the *updated* observer $\phi^{(k+1)}$. This discrepancy is defined as the **Adaptation Lag**.

**Theorem 2** (Meta-Gradient Anticipation). *The meta-level update in DeMP incorporates a second-order gradient term that explicitly minimizes the sensitivity of the policy to the observer's learning dynamics, thereby mitigating adaptation lag.*

*Proof.* Consider the meta-objective $\mathcal{L}^{(\text{meta})}(\theta^{(0)})$ over a simplified sequence of length $M = 1$ for clarity. The meta-loss measures the performance of the adapted policy $\theta^{(1)}$ against the *updated* observer $\phi^{(1)}$:

$$\mathcal{L}^{(\text{meta})} = \mathcal{L}(\theta^{(1)}, \phi^{(1)}),$$

where $\theta^{(1)} = \theta^{(0)} - \alpha \nabla_\theta \mathcal{L}(\theta^{(0)}, \phi^{(0)})$ (agent adaptation) and $\phi^{(1)} = \phi^{(0)} - \eta \nabla_\phi \mathcal{L}_{\mathcal{O}}(\phi^{(0)}, \tau(\theta^{(0)}))$ (observer adaptation).

We compute the gradient of the meta-objective with respect to the initialization $\theta^{(0)}$:

$$\nabla_{\theta^{(0)}} \mathcal{L}^{(\text{meta})} = \underbrace{\frac{\partial \mathcal{L}}{\partial \theta^{(1)}} \frac{\partial \theta^{(1)}}{\partial \theta^{(0)}}}_{\text{Direct Policy Optimization}} + \underbrace{\frac{\partial \mathcal{L}}{\partial \phi^{(1)}} \frac{\partial \phi^{(1)}}{\partial \theta^{(0)}}}_{\text{Observer Anticipation Term}}.$$

The first term corresponds to standard model-agnostic meta-learning (MAML), finding an initialization that adapts fast. The second term is unique to the RDPP setting. Let us expand $\frac{\partial \phi^{(1)}}{\partial \theta^{(0)}}$ using the observer's update rule:

$$\frac{\partial \phi^{(1)}}{\partial \theta^{(0)}} = -\eta \frac{\partial}{\partial \theta^{(0)}} \left( \nabla_\phi \mathcal{L}_{\mathcal{O}}(\phi^{(0)}, \tau) \right) = -\eta \nabla^2_{\phi,\tau} \mathcal{L}_{\mathcal{O}} \cdot \frac{\partial \tau}{\partial \theta^{(0)}}.$$

Substituting this back, the meta-gradient contains a term proportional to:

$$-\eta \left( \nabla_{\phi^{(1)}} \mathcal{L} \right)^T \left( \nabla^2_{\phi,\tau} \mathcal{L}_{\mathcal{O}} \right) \frac{\partial \tau}{\partial \theta^{(0)}}.$$

**Interpretation:** This term represents the gradient of the loss with respect to how the agent's trajectory ($\tau$) interacts with the observer's learning process ($\nabla \mathcal{L}_{\mathcal{O}}$). By descending this gradient, DeMP updates $\theta^{(0)}$ to minimize not just the current loss, but the expected future loss after the observer's update $\phi^{(0)} \to \phi^{(1)}$. In essence, the agent achieves proactive adaptation: instead of merely reacting to the current model, it anticipates the shift in the observer's decision boundary and identifies policy parameters that remain deceptive despite this shift. This explicit anticipation bridges the temporal gap between the optimization step and the evaluation step, thereby mitigating the adaptation lag. □

# B Environment Setting

## B.1 Environment Details

Following the environment setup in DEAM Lewis and Miller (2023), we adopt a discrete two-dimensional (2D) grid-world adapted from the MiniGrid framework Chevalier-Boisvert et al. (2023). This environment allows direct comparison with prior deceptive path planning studies while maintaining computational simplicity. It consists of an agent navigating from a start position to its true goal while avoiding obstacles and misleading an external observer. We extend the original MiniGrid configuration to support diagonal movements and distance-based step penalties instead of binary completion rewards. Multiple candidate goals are embedded within a single map instance, enabling probabilistic goal inference by the observer during interaction.

**State Space ($\mathcal{S}$).** At each timestep, the agent receives its current spatial coordinate as the state. Each state $s_t \in \mathcal{S}$ is represented as a two-dimensional tuple $(x_t, y_t)$ denoting the agent's position on the grid. We evaluate our method on grid maps of two different scales: $49 \times 49$ and $100 \times 100$. In both configurations, obstacles occupy a subset of cells that are impassable. The agent's field of view covers the full map, and state transitions are deterministic.

**Action Space ($\mathcal{A}$).** The agent can move in eight discrete directions:

$$\mathcal{A} = \{\text{up}, \text{down}, \text{left}, \text{right}, \text{up-left}, \text{up-right}, \text{down-left}, \text{down-right}\}.$$

Each action changes the position by one grid cell in the corresponding direction, unless the next cell is blocked by an obstacle, in which case the agent remains stationary. This yields a low-dimensional deterministic action space suitable for trajectory-level reasoning and meta-adaptive updates.

**Reward Function ($\mathcal{R}$).** At each timestep, the environment provides a scalar reward according to the agent's movement and goal-reaching behavior:

$$r_t = \begin{cases} +100, & \text{if the agent reaches the true goal,} \\ -1, & \text{if the agent moves horizontally or vertically,} \\ -\sqrt{2}, & \text{if the agent moves diagonally.} \end{cases}$$

This distance-based reward encourages efficient trajectories while preserving the deceptive planning objective. Episodes terminate when the agent reaches any candidate goal or after 150 timesteps.

**Map Configurations.** All experiments are conducted on $49 \times 49$ grid maps under two configurations: (1) maps with large structured obstacles forming corridors and multiple goal regions; and (2) maps with randomly distributed high-density obstacles. Each map contains five candidate goals, and the true goal is randomly selected from this set at the beginning of each episode. Crucially, each of the two map configurations is evaluated using five different goal/start-state configurations. Across 100 random trials, one of these five specific goal/start-state pairs is randomly selected for the beginning of the experiment.

## B.2 Observer Settings

Following the experimental protocol described in the main text, the observer is implemented as a supervised sequence model that performs goal-recognition from partial agent trajectories. The observer maps an

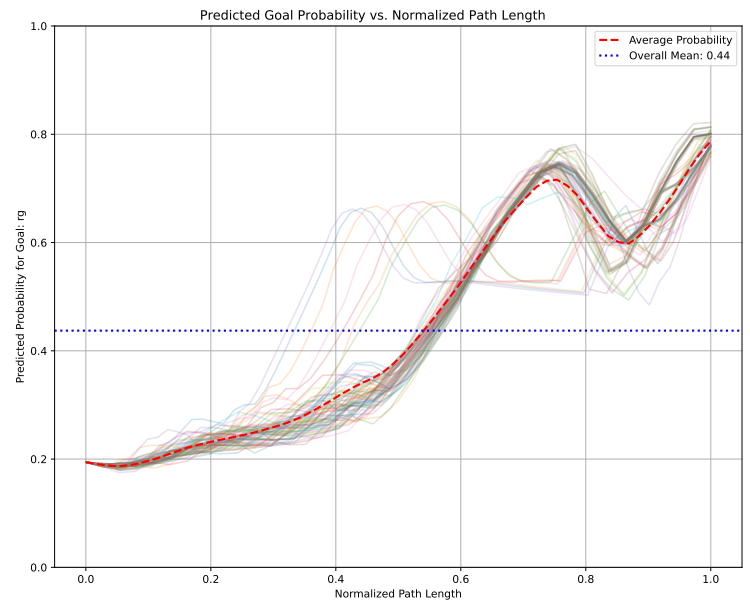


(a) Observer's prediction in pre-training 200 episodes

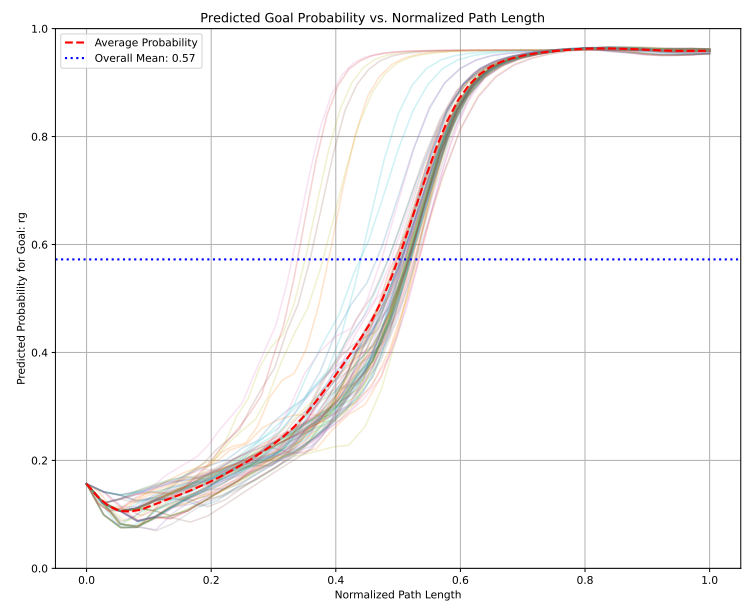


(b) Observer's prediction in pre-training 400 episodes

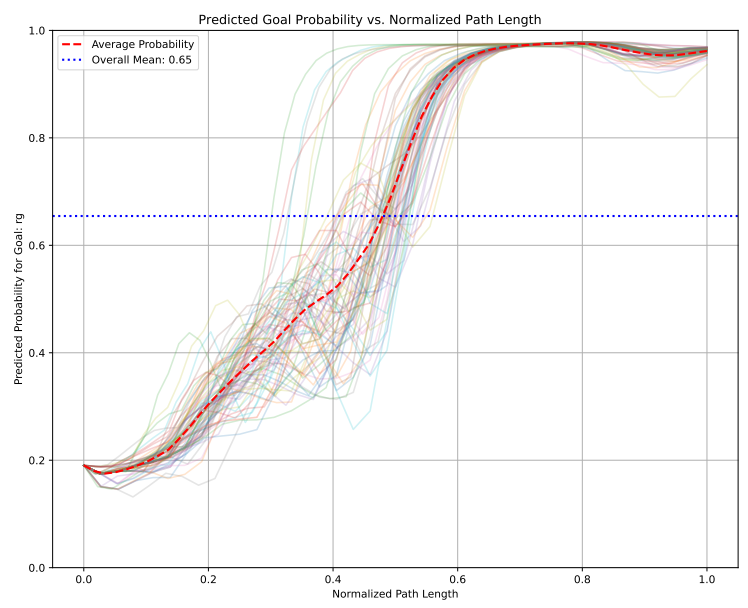


(c) Observer's prediction in pre-training 800 episodes

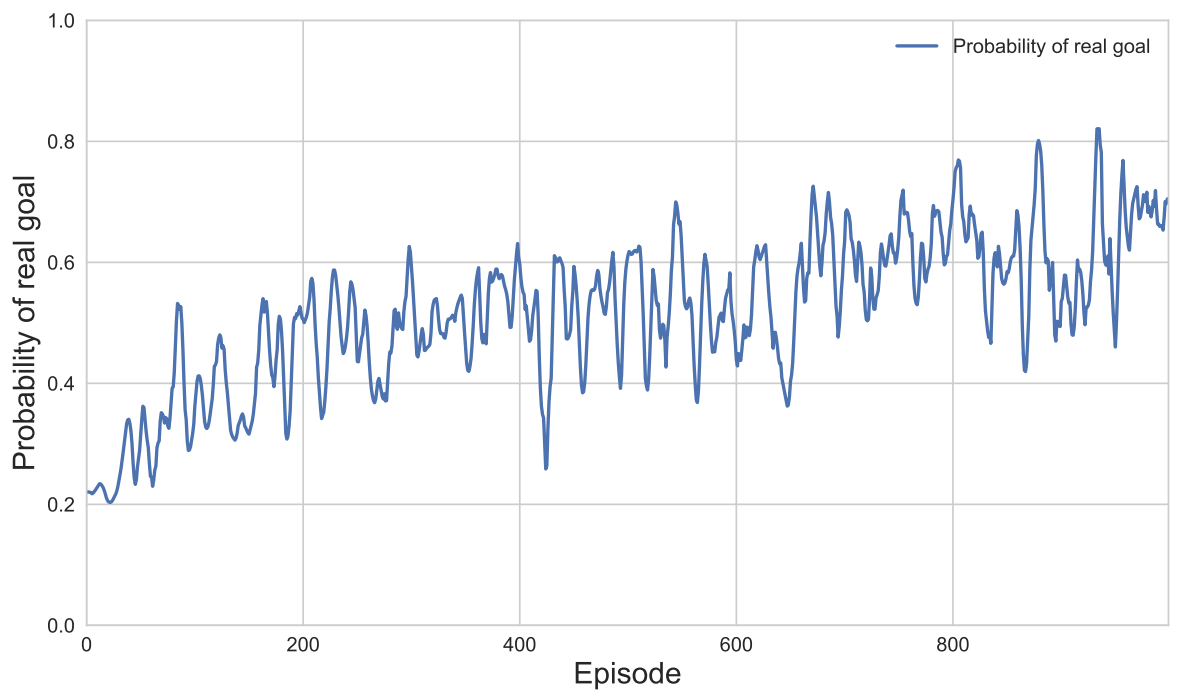


(d) Observer's prediction of real goal across 1000 pre-training episodes

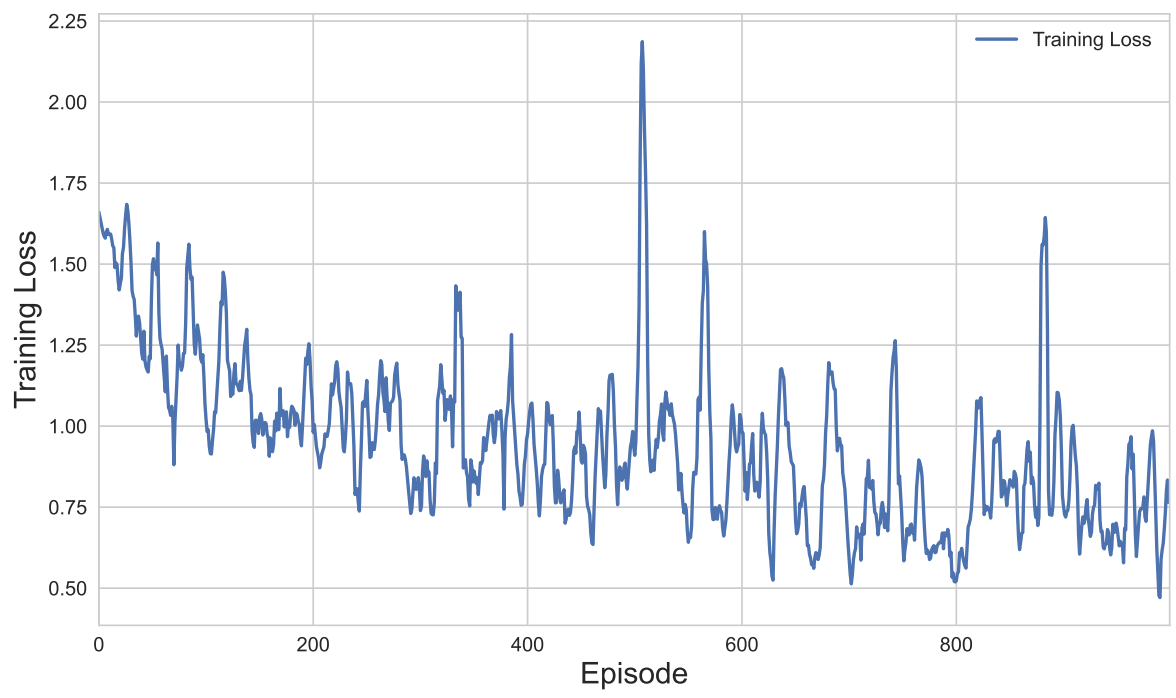


(e) Observer's training loss

Figure 1: Observer pretraining dynamics and predictive performance. (a–c) Predicted probability of the true goal along normalized trajectories generated by a non-deceptive planner, evaluated at different stages of observer pretraining (200, 400, and 800 episodes). (d) Evolution of the observer's predicted probability for the true goal across 1000 pretraining episodes. (e) Corresponding training loss during pretraining. The results show that pretraining enables the observer to develop consistent goal-recognition behavior prior to deceptive interaction.

observed trajectory prefix to a probability distribution over the candidate goal set $\mathcal{G}$ (size $|\mathcal{G}| = 5$ in our experiments). Below we detail the model architecture, input representation, the offline pre-training procedure used to establish baseline recognition ability, and the online fine-tuning protocol used during repeated interactions.

**Model Architecture.** The observer model is implemented as a two-layer Long Short-Term Memory (LSTM) network that encodes the agent's movement trajectory. At each timestep $t$, the input is the agent's spatial coordinate $(x_t, y_t)$, and the output is a five-dimensional vector representing unnormalized goal likelihoods. Formally, given a trajectory $\tau = \{(x_1, y_1), \ldots, (x_T, y_T)\}$, the LSTM processes the padded sequence and extracts the hidden representation of the last valid timestep, which is passed through a linear projection layer to yield the final goal logits. No additional normalization or attention layers are applied, ensuring the observer's prediction depends purely on trajectory dynamics.

**Input Representation and Data Augmentation.** The observer conditions on the sequence of agent coordinates only and agent actions are not used. Candidate goal coordinates and the map geometry are available through the training data distribution. To improve robustness to partial observations, during pre-training we

sample prefixes of varying relative lengths: for each full trajectory we uniformly sample a prefix length ratio $\rho \in [0.1, 1.0]$ and use the first $\lceil \rho T \rceil$ timesteps as the input. This forces the observer to learn goal-recognition at different observation completeness levels.

**Pretraining Procedure.** The observer is pretrained prior to any interaction with the deceptive agent to approximate the observer's initial predictive prior. Training data consist of trajectories generated by a non-deceptive shortest-path planner across all map configurations and goal/start-state pairs. The pretraining loss is a standard classification objective over the true goal labels, optimized until convergence.

Figure 1 visualizes the observer's predictive behavior during pretraining. As shown in Figure 1 (a–c), as pretraining progresses, the observer increasingly assigns higher confidence to the true goal earlier along the trajectory, indicating improved goal-recognition capability. Figure 1 (d) shows the evolution of the predicted probability of the true goal over the entire pretraining process, while Figure 1 (e) reports the corresponding training loss. Together, these results confirm that pretraining yields a stable and informative initialization for the observer before entering the repeated deceptive interaction phase.

**Online Finetune.** During online interaction, the observer first receives a partial trajectory prefix $\tau_{1:t}^{\text{obs}}$ whose relative length is uniformly sampled from $[0.4, 0.6]$. Based on this partial observation, it outputs a predictive distribution $\hat{p}(g \mid \tau_{1:t}^{\text{obs}})$ over all candidate goals, which is then fed back to the agent as the observer's signal for adaptive deceptive planning. After the episode completes, the observer obtains access to the full trajectory $\tau_{1:T}$ along with the true goal label and performs an online fine-tuning step. A single batch gradient update is applied using this most recent trajectory batch, augmented in the same manner as during pretraining. This two-stage design allows the observer to continually refine its inference model based on new experience while remaining a bounded, learnable opponent.

# C Implementation Details

## C.1 DeMP Implementation and Training

DeMP is implemented on top of the Soft Actor-Critic (SAC) algorithm Haarnoja et al. (2018). Both the policy and value networks are two-layer multilayer perceptrons (MLPs) with 256 hidden units per layer. The agent is first initialized via imitation learning on trajectories generated by the AM method Liu et al. (2021) (behavioral cloning), and then trained online under the RDPP protocol using a two-level optimization procedure: an *episode-level* adaptation after each episode and a *meta-level* update every $M = 2$ episodes.

To avoid a notation clash with SAC's entropy temperature, we denote the episode-level learning rate by $\alpha$ (set to $1\text{e}{-3}$ in our experiments) and the meta-level learning rate by $\beta$ (set to $1\text{e}{-4}$). The SAC entropy temperature is adapted automatically during training. Unless stated otherwise, all remaining SAC hyperparameters follow the defaults in Haarnoja et al. Haarnoja et al. (2018).

**Imitation pretraining (behavioral cloning).** To provide a strong deceptive initialization, the agent is pretrained by behavioral cloning on expert-like trajectories produced by the AM method Liu et al. (2021). We collect a large dataset of AM trajectories across the evaluation maps ($1 \times 10^4$ trajectories). At each timestep the agent state is paired with the AM action; the policy is trained with a supervised loss to match these actions. Imitation learning is performed with the Adam optimizer (learning rate $1\text{e}{-3}$, batch size 128) and training is continued until validation performance plateaus. This pretrained policy serves as the first episode initialization $\theta^{(0)}$ for subsequent DeMP online adaptation.

**Meta-level update.** After every $M$ episodes we perform a meta-update over the initialization parameters $\theta^{(0)}$ to accelerate future adaptation. In practice, to reduce computation, we adopt a first-order approximation, i.e., we approximate the meta-gradient by ignoring second-order derivative terms resulting from the dependence of $\theta^{(M)}$ on $\theta^{(0)}$. In practice this yields stable meta-updates with substantially lower memory and compute cost while retaining the desired fast-adaptation behavior.

## C.2 Practical Details and Notes

- **Episode length and evaluation.** Episodes terminate when the agent reaches any candidate goal or after $150$ timesteps. Each evaluation sequence consists of $400$ online episodes, and reported results are averaged across $100$ independent random seeds (repetitions).
- **Observer feedback.** In each episode the observer returns a predictive distribution computed from a partial trajectory prefix whose relative length is uniformly sampled from $[0.4, 0.6]$; this distribution is available to the agent during its episode-level adaptation.
- **Replay buffer and batch updates.** For the episode-level adaptation we explicitly use data from the current episode (single batch gradient update) as described above.

## C.3 DeMP Hyperparameters

Other SAC hyperparameters are left at the standard defaults from Haarnoja et al. Haarnoja et al. (2018) unless otherwise specified above.

Table 1: DeMP hyperparameters

| Parameter | Symbol | Value |
|---|---|---|
| Policy / value network | – | 2-layer MLP, 256 hidden units per layer |
| Episode-level learning rate | $\alpha$ | $1e{-3}$ |
| Meta-level learning rate | $\beta$ | $1e{-4}$ |
| Meta update interval | $M$ | 2 episodes |
| SAC entropy temperature | – | adaptive (auto-tuned) |
| SAC batch size | – | 128 |
| Replay buffer capacity | – | $10^5$ transitions |
| Discount factor | $\gamma$ | 0.99 |
| Max steps per episode | – | 150 |
| Imitation pretraining dataset | – | $\approx$ 10000 AM trajectories |
| Imitation pretraining optimizer | – | Adam, lr $1e{-3}$, batch size 128 |
| Evaluation episodes per run | – | 400 |
| Independent runs (random seeds) | – | 100 |



\end{document}